\documentclass[12pt]{article}
\usepackage{enumerate,natbib,mathrsfs,amsthm,amsmath,bm,amssymb,listings,setspace}
\usepackage[OT1]{fontenc}
\usepackage[colorlinks,linkcolor=red,anchorcolor=blue,citecolor=blue]{hyperref}
\usepackage{fullpage}
\usepackage[protrusion=false,expansion=true]{microtype}
\usepackage{algorithm,algorithmic,graphicx}

\theoremstyle{plain}
\let\hat\widehat
\let\tilde\widetilde

\newcommand{\ab}{\mathbf{a}}
\newcommand{\bbb}{\mathbf{b}}
\newcommand{\cbb}{\mathbf{c}}

\newcommand{\ub}{\mathbf{u}}
\newcommand{\vb}{\mathbf{v}}
\newcommand{\wb}{\mathbf{w}}

\newcommand{\yb}{\mathbf{y}}

\newcommand{\ba}{\bm{a}}

\newcommand{\bu}{\bm{u}}
\newcommand{\bv}{\bm{v}}
\newcommand{\bw}{\bm{w}}

\newcommand{\Bb}{\mathbf{B}}

\newcommand{\Db}{\mathbf{D}}

\newcommand{\Ib}{\mathbf{I}}

\newcommand{\bI}{\bm{I}}

\newcommand{\cA}{\mathcal{A}}
\newcommand{\cB}{\mathcal{B}}

\newcommand{\cE}{\mathcal{E}}
\newcommand{\cF}{\mathcal{F}}
\newcommand{\cG}{\mathcal{G}}

\newcommand{\cM}{\mathcal{M}}

\newcommand{\cS}{{\mathcal{S}}}
\newcommand{\cT}{{\mathcal{T}}}

\newcommand{\cX}{\mathcal{X}}

\newcommand{\bbeta}{\bm{\beta}}

\newcommand{\bepsilon}{\bm{\epsilon}}

\newcommand{\bmu}{\bm{\mu}}

\newcommand{\bSigma}{\bm{\Sigma}}

\DeclareMathOperator{\ind}{\mathrm{1}}  

\newtheorem{lemma}{{\bf Lemma}}

\newtheorem{corollary}{{\bf Corollary}}

\newtheorem{theorem}{{\bf Theorem}}

\newtheorem{assumption}{{\bf Assumption}}

\def\eop
{\hfill $\Box$ \\

}

\def\change{\textcolor{black}}

\newcommand{\Rbb}{{\mathbb R}}

\title{\Large{\textbf{Dynamic Tensor Clustering}}}
\author{
\bigskip
Will Wei Sun and Lexin Li \\ 
\normalsize{University of Miami and University of California at Berkeley}
}
\date{}

\begin{document}

\maketitle

\begin{footnotetext}[1]
{Will Wei Sun is Assistant Professor, Department of Management Science, University of Miami Business School, Miami, FL 33146. Email: wsun@bus.miami.edu. Sun's work was partially supported by Provost's research award from University of Miami. Lexin Li is Professor, Division of Biostatistics, University of California, Berkeley, Berkeley, CA 94720-3370. Email: lexinli@berkeley.edu. Li's work was partially supported by NSF grant DMS-1613137 and NIH grant AG034570. Both authors thank the editor, one associate editor and three reviewers for their helpful comments and suggestions which led to a much improved presentation.}
\end{footnotetext}

\begin{abstract}
Dynamic tensor data are becoming prevalent in numerous applications. Existing tensor clustering methods either fail to account for the dynamic nature of the data, or are inapplicable to a general-order tensor. There is also a gap between statistical guarantee and computational efficiency for existing tensor clustering solutions. In this article, we propose a new dynamic tensor clustering method that works for a general-order dynamic tensor, and enjoys both strong statistical guarantee and high computational efficiency. Our proposal is based on a new structured tensor factorization that encourages both sparsity and smoothness in parameters along the specified tensor modes. Computationally, we develop a highly efficient optimization algorithm that benefits from substantial dimension reduction. Theoretically, we first establish a non-asymptotic error bound for the estimator from the structured tensor factorization. Built upon this error bound, we then derive the rate of convergence of the estimated cluster centers, and show that the estimated clusters recover the true cluster structures with high probability. Moreover, our proposed method can be naturally extended to co-clustering of multiple modes of the tensor data. The efficacy of our method is illustrated through simulations and a brain dynamic functional connectivity analysis from an Autism spectrum disorder study. 
\end{abstract}


\newpage
\baselineskip=19pt

\section{Introduction}
\label{sec:introduction}

Data in the form of multidimensional array, or tensor, are now frequently arising in a diverse range of scientific and business applications \citep{zhu2007statistical, liu2013, zhou2013, ding2015tensor}. Particularly, for a large class of tensor data, time is one of the tensor modes, and this class is often termed as \emph{dynamic tensor}. Examples of dynamic tensor data are becoming ubiquitous and its analysis are receiving increasing attention. For instance, in online advertising, \cite{bruce2016} studied consumer engagement on advertisements over time to capture the temporal dynamics of users behavior. The resulting data is a three-mode tensor of user by advertisement by time. In molecular biology, \citet{Seigal2016} studied time-course measurements of the activation levels of multiple pathways from genetically diverse breast cancer cell lines after exposure to numerous growth factors with different dose. The data is a five-mode tensor of cell line by growth factor by pathway by dose by time. In neurogenomics, \citet{LiuYuan2017} modeled spatial temporal patterns of gene expression during brain development, and the data is a three-mode tensor of gene by brain region by time. In Section \ref{sec:realdata}, we illustrate our method on a brain dynamic connectivity analysis, where the goal is to understand interaction of distinct brain regions and their dynamic pattern over time. One form of the data in this context is a three-mode tensor of region by region by time. 

Clustering has proven to be a useful tool to reveal important underlying data structures \citep{yuan2006, ma2008, shen2012clustering, wang2013provable}. Directly applying a clustering algorithm to the vectorized tensor data is a simple solution, but it often suffers from poor clustering accuracy, and induces heavy and sometimes intractable computation. There have been a number of proposals for clustering of tensor data, or the two-mode special case, matrix data. One class of such methods focused on biclustering that simultaneously group rows (observations) and columns (features) of the data matrix \citep{huang2009, lee2010, chi2015, chi2016}. The proposed solutions were based on sparse singular value decomposition, or a reformulation of biclustering as a penalized regression with some convex penalties. However, the dynamic property remained largely untapped in those solutions. The second class of approaches directly targeted biclustering of time-varying matrix data \citep{hocking2011, ji2012, li2015}. Nevertheless, those solutions were designed specifically for matrix-valued data. Extension to a general-order tensor is far from trivial. The third class tackled clustering of tensor data through some forms of $\ell_1$ penalization \citep{cao2013, wu2016}. But none of those approaches incorporated the dynamic information in the data and would inevitably lead to loss in clustering accuracy. Moreover, there is no statistical guarantee provided in the performance of these tensor clustering algorithms. \emph{Tensor decomposition} is a crucial component in handling tensor-valued data, and is to play a central role in our proposed tensor clustering solution as well. There have been a number of recent proposals of convex relaxation of tensor decomposition through various norms \citep{romera2013, yuanzhang2016, yuan2016, zhang2016cross}. However, all those methods focused on low-rank tensor recovery, and none incorporated any sparsity or fusion structure in the decomposition. Recently, \citet{sun2016provable} proposed a low-rank decomposition with a truncation operator for hard thresholding. Although sparsity was considered, they did not consider fusion structure for a dynamic tensor. Ignoring this structure, as we show later, would induce large estimation errors in both tensor decomposition and subsequent clustering. In summary, there exists no clustering solution with statistical guarantee that handles a general-order dynamic tensor and incorporates both sparsity and fusion structures. 

In this article, we aim to bridge this gap by proposing a dynamic tensor clustering method, which takes into account both sparsity and fusion structures, and enjoys strong statistical guarantee as well as high computational efficiency. Our proposal makes multiple contributions to the clustering literature. First, our clustering method is built upon a newly proposed structured tensor factorization approach, which encourages both sparsity and smoothness in the decomposed components, and in turn captures the dynamic nature of the tensor data. We show how structured tensor factorization can be used to infer the clustering structure. Interestingly, we find that tensor Gaussian mixture model can be viewed as a special case of our clustering method. Second, our proposal is computationally efficient. This is partly achieved by substantial dimension reduction resulting from the imposed structure; in the illustrative example in Section \ref{sec:realdata}, the number of free parameters was reduced from about two millions to one thousand. Our optimization algorithm can be decomposed as an unconstrained tensor decomposition step followed by a constrained optimization step. We show that, the overall computational complexity of our constrained solution is comparable to that of the unconstrained one. Third, and probably most importantly, we establish rigorous theoretical guarantee for our proposed dynamic tensor clustering solution. Specifically, we first establish a non-asymptotic error bound for the estimator from the proposed structured tensor factorization. Based on this error bound, we then obtain the rate of convergence of the estimated cluster centers from our dynamic tensor clustering solution, and prove that the estimated clusters recover the true cluster structures with high probability. It is also noteworthy that we allow the number of clusters to grow with the sample size. Such consistency results are new in the tensor clustering literature. From a technical perspective, the fusion structure we consider introduces some additional challenges in the theoretical analysis, since the resulting  truncated fusion operator is non-convex and the observed tensor is usually noisy with an unknown error distribution. To address such challenges, we develop a set of non-asymptotic techniques to carefully evaluate the estimation error in each iteration of our alternating updating algorithm. We also utilize a series of large deviation bounds to show that our estimation error replies on the error term only through its sparse spectral norm, which largely relieves the uncertainty from the unknown error distribution. Last but not least, although our algorithm mostly focuses on clustering along a single mode of tensor data, the same approach can be easily applied to co-clustering along multiple tensor modes. This is different from classical clustering methods, where an extension from clustering to bi-clustering generally requires different optimization formulations \citep[see, e.g.,][]{chi2016}. In contrast, our clustering method naturally incorporates single and multi-mode clustering without requiring any additional modification.

The rest of the article is organized as follows. Section \ref{sec:DTC} introduces the proposed dynamic tensor clustering method, and Section \ref{sec:STF} presents its solution through structured tensor factorization. Section \ref{sec:theorem} establishes the estimation error bound of the structured tensor factorization and the consistency properties of dynamic tensor clustering. Section \ref{sec:simulations} presents the simulations, and Section \ref{sec:realdata} illustrates with a brain dynamic functional connectivity analysis. The Supplementary Materials collect all technical proofs and additional simulations.

\section{Model}
\label{sec:DTC}

\subsection{Clustering via tensor factorization}
\label{sec:cluster}

Given $N$ copies of $m$-way tensors, $\cX_1, \ldots, \cX_N \in \Rbb^{d_1 \times \cdots \times d_m}$, our goal is to uncover the underlying cluster structures of these $N$ samples. That is, we seek the true cluster assignment, 
\begin{eqnarray*}
(\underbrace{1, \ldots, 1}_{l \textrm{~samples}}, \; \underbrace{2, \ldots, 2}_{l \textrm{~samples}}, \; \ldots, \; \underbrace{K, \ldots, K}_{l \textrm{~samples}}),
\end{eqnarray*}
where $K$ is the number of clusters and $l = N/K$. Here, for ease of presentation, we assume an equal number of $l$ samples per cluster. \change{In Section \ref{sec:additional_experiments} of the Supplementary Materials, we report some numerical results with unequal cluster sizes.}

To cluster those tensor samples, we first stack them into a $(m+1)$-way tensor, $\cT \in \Rbb^{d_1 \times \cdots \times d_m \times N}$. We comment that, in principle, one can cluster along any single or multiple modes of $\cT$. Without loss of generality, we focus our discussion on clustering along the last mode of $\cT$, and only briefly comment on the scenario that clusters along multiple modes. This formulation covers a variety of scenarios encountered in our illustrative example of brain dynamic connectivity analysis. For instance, in one scenario, $\cT \in \Rbb^{p \times p \times t \times n}$, $N = n$, and the goal is to cluster $n$ individuals, each with a $p \times p \times t$ tensor that represents the brain connectivity pattern among $p$ brain regions over $t$ sliding time windows. In another scenario, $\cT \in \Rbb^{p \times p \times t}$, $N = t$, and the goal is to cluster $t$ sliding windows for a single subject. In the third scenario, $\cT \in \Rbb^{p \times p \times n_g \times t}$, $N = t$, where $n_g$ is the number of subjects in group $g$, and the goal becomes clustering $t$ moving windows for all subjects in that group. 

Our key idea is to consider a structured decomposition of $\cT$, then apply a usual clustering algorithm, e.g., $K$-means, to the matrix from the decomposition that corresponds to the last mode to obtain the cluster assignment. Assume that the tensor $\cT$ is observed with noise, 
\begin{eqnarray} \label{eqn:additive-model} 
\cT = \cT^* + \cE, 
\end{eqnarray}
where $\cE$ is an error tensor, and $\cT^*$ is the true tensor with a rank-$R$ CANDECOMP/PARAFAC (CP) decomposition structure \citep{kolda2009}, 
\begin{equation} \label{eqn:CP}
\cT^* =  \sum_{r=1}^R w^*_r \bbeta^*_{1,r} \circ \cdots \circ \bbeta^*_{m+1,r},
\end{equation}
where $\bbeta_{j,r}^* \in \mathbb R^{d_j}$, $\|\bbeta_{j,r}^*\|_2 = 1$, $w^*_r > 0$, $j=1, \ldots, m+1, r=1,\ldots,R$, $\| \cdot \|_2$ denotes the vector $\ell_2$ norm and $\circ$ is the vector outer product. For ease of notation, we define $d_{m+1} := N$. \change{We have chosen the CP decomposition due to its relatively simple formulation and its competitive empirical  performance. It has been widely used for link prediction \citep{dunlavy2011}, community detection \citep{anandkumar2014a}, recommendation systems \citep{bi2017}, and convolutional neural network speeding-up \citep{lebedev2015}.}

\begin{figure}[t!]
\centering
\vskip 0.1em
\includegraphics[scale=0.43]{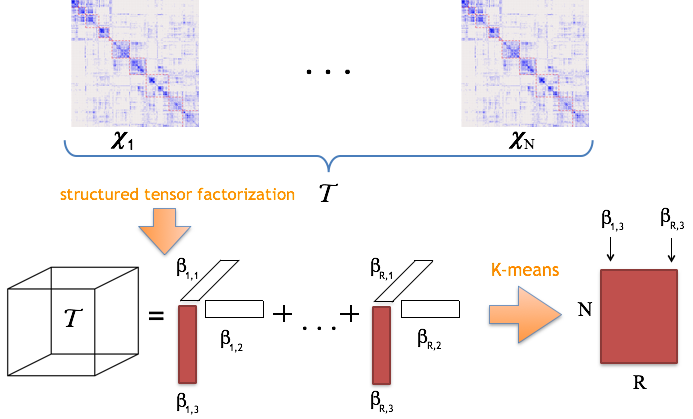}
\caption{A schematic illustration of the proposed tensor clustering method. It stacks multiple tensor samples to a higher-order tensor, carries out structured tensor factorization, then applies a classical clustering algorithm, e.g, $K$-means, to the data matrix from the decomposition that corresponds to the last mode of the stacked tensor.}
\label{fig:stack_tensor}
\end{figure}

Given the structure in \eqref{eqn:CP}, it is straightforward to see that, the cluster structure of samples along the last mode of the tensor $\cT$ is fully determined by the matrix that stacks the decomposition components, $\bbeta^*_{m+1,1}, \ldots, \bbeta^*_{m+1,R}$. We denote this matrix as $\Bb_{m+1}^*$, which can be written as 
\begin{eqnarray}
\Bb_{m+1}^* := \left( \bbeta^*_{m+1,1}, \ldots, \bbeta^*_{m+1,R} \right) = \Bigl( \underbrace{\bmu_{1}^{* \top}, \ldots, \bmu_{1}^{*\top}}_{l \textrm{~samples}}, \ldots, \underbrace{\bmu_{K}^{*\top}, \ldots, \bmu_{K}^{*\top}}_{l \textrm{~samples}} \Bigr)^{\top} \in \Rbb^{N \times R},
\label{eqn:true_cluster_center}
\end{eqnarray}
where $\bmu^*_{k} := (\mu^*_{k,1}, \ldots, \mu^*_{k,R}) \in \Rbb^{R}$, $k=1,\ldots,K$, indicates the cluster assignment. Figure \ref{fig:stack_tensor} shows a schematic illustration of our tensor clustering proposal. We comment that, if the goal is to cluster along more than one tensor mode, one only needs to apply a clustering algorithm to the matrix formed by each of those modes separately.

Accordingly, the true cluster means of the tensor samples $\cX_1, \ldots, \cX_N$ can be written as, 
\begin{eqnarray} \label{eqn:clustermean}
\underbrace{\cM_1 := \sum_{r=1}^R w^*_r \bbeta^*_{1, r} \circ \cdots \circ \bbeta^*_{m, r} \mu^*_{1,r}}_{\textrm{cluster center 1}}, \quad \ldots, \quad \underbrace{\cM_K :=\sum_{r=1}^R w^*_r \bbeta^*_{1, r} \circ \cdots \circ \bbeta^*_{m, r} \mu^*_{K,r}}_{\textrm{cluster center K}}.
\end{eqnarray}
The structure in \eqref{eqn:clustermean} reveals the key underlying assumption of our tensor clustering solution. That is, we assume each cluster mean is a linear combination of the outer product of $R$ rank-1 basis tensors, and all the cluster means share the same $R$ basis tensors. We recognize that this assumption introduces an additional constraint. However, it leads to substantial dimension reduction, which in turn enables efficient estimation and inference in subsequent analysis. As we show in Section \ref{sec:TGMM}, the tensor Gaussian mixture model can be viewed as a special case of our clustering structure. Moreover, as our numerical study has found, this imposed structure provides a reasonable approximation in real data applications. 

\begin{figure}[t!]
\centering
\includegraphics[scale=0.35]{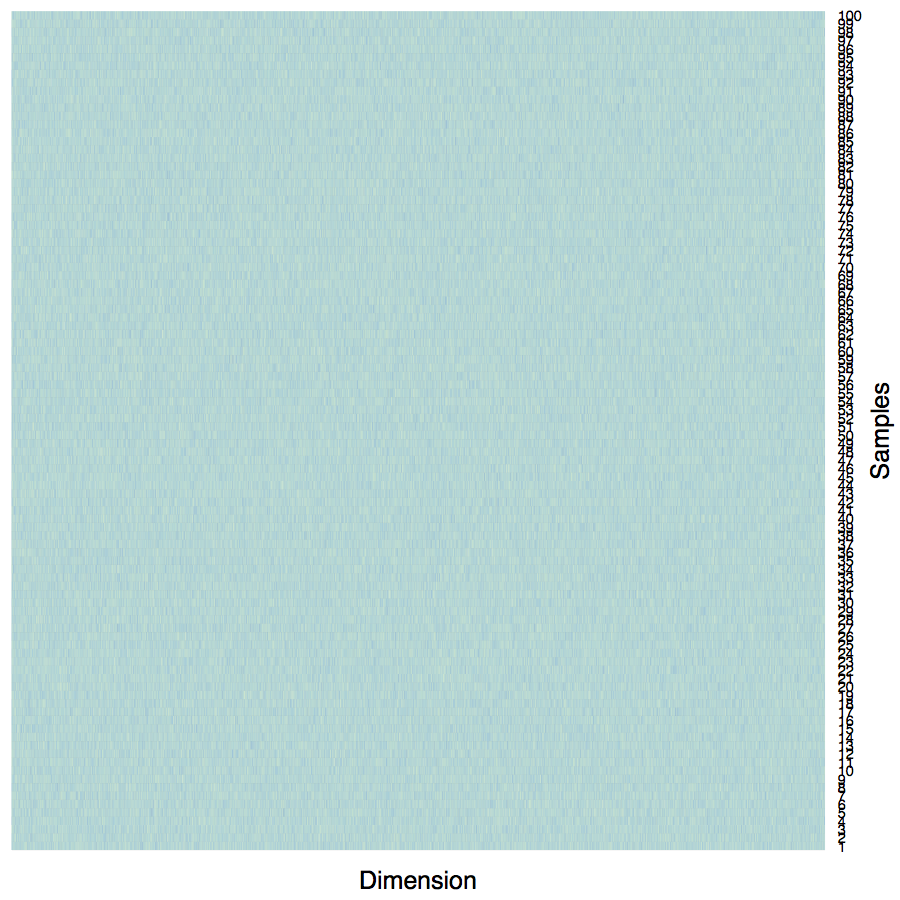}~~
\includegraphics[scale=0.35]{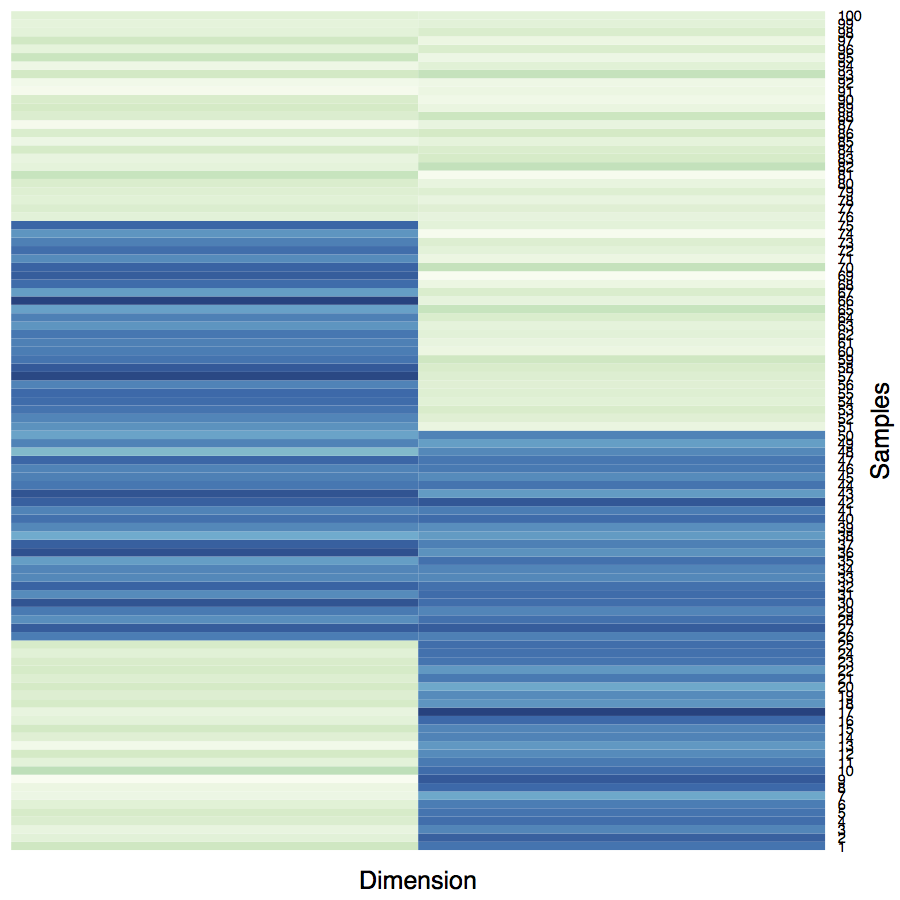}
\caption{Heatmap of the vectorized data, with dimension $100 \times 8000$, is shown in the left panel, and heatmap of the reduced data from our tensor factorization, with dimension $100 \times 2$, is shown on the right. The true cluster structure can be fully recovered by any reasonable clustering method based on the reduced data, but not based on the vectorized data.}
\label{fig:heatmap}
\end{figure}

For comparison, we consider the alternative solution that applies clustering directly on the vectorized version of tensor data. It does not require \eqref{eqn:clustermean}, and the corresponding number of free parameters is in the order of $K\prod_j d_j$. In the example in Section~\ref{sec:realdata}, $d_1=d_2=116, d_3=80, K=2$, and that amounts to $2,152,960$ parameters. Imposing \eqref{eqn:clustermean}, however, would reduce the number of free parameters to $R(\sum_j {d_j} + K + 1)$; again, for the aforementioned example, $R=5$ and that amounts to $1,175$ parameters. Such a substantial reduction in dimensionality is crucial for both computation and inference in tensor data analysis. Moreover, we use a simple simulation example to demonstrate that, our clustering method, which assumes \eqref{eqn:clustermean} and thus exploits the underlying structure of the tensor data, can not only reduce the dimensionality and the computational cost, but also improve the clustering accuracy. Specifically, we follow the example in Section \ref{sec:sim-2d} to generate $N =100$ tensor samples of dimension $d_1 = d_2 = d_3 = 20$ from 4 clusters, with samples 1 to 25, 26 to 50, 51 to 75, 76 to 100 belonging to clusters 1 to 4, respectively. Figure \ref{fig:heatmap} shows the heatmap of the vectorized data (left panel), which is of dimension $100 \times 8000$, and the heatmap of the data with reduced rank $R=2$ (right panel), which is of dimension $100\times 2$. It is clearly seen from this plot that, our clustering method based on the reduced data under \eqref{eqn:clustermean} is able to fully recover the four underlying clusters, while the clustering method based on the vectorized data cannot.

\subsection{Sparsity and fusion structures}
\label{sec:stf_model}

Motivated from the brain dynamic functional connectivity analysis, in addition to the CP low-rank structure \eqref{eqn:CP}, we also impose the sparsity and smoothness fusion structures in tensor decomposition to capture the sparsity and dynamic properties of the tensor samples.  
Specifically, we impose the following structures in the parameter space,
\begin{eqnarray*}
\cS(d, s_0) & := & \left\{ \bbeta \in \mathbb R^d \Big | \sum_{j=1}^{d} \ind_{\{\beta_{j}  \ne 0\}} \le s_0 \right\}, \\
\cF(d, f_0) & := & \left\{ \bbeta \in \mathbb R^d \Big |  \sum_{j=2}^{d} \left |\beta_{j} - \beta_{j-1} \right| \le f_0 \right\} = \left\{ \bbeta \in \mathbb R^d \Big|  \left \|  \Db \bbeta \right \|_1 \le f_0 \right\},
\end{eqnarray*}
where $\bbeta = (\beta_1, \ldots, \beta_d)^{\top}$, $\|\cdot\|_1$ denotes the vector $\ell_1$ norm, and $\Db \in \mathbb R^{(d-1) \times d}$, whose $j$th row has $-1$ and $1$ on its $j$th and $(j+1)$th positions and zero elsewhere. 
Combining these two structures with the CP decomposition of $\cT^*$ in \eqref{eqn:CP}, we consider 
\begin{eqnarray*}
\bbeta_{j,r}^* \in \cS(d_j, s_{0,j}) \cap \cF(d_j, f_{0,j}), \textrm{~for~any~} j=1,\ldots, m+1, \; r=1,\ldots,R.
\end{eqnarray*}
Here for simplicity, we assume the maximum of the sparsity parameter $s_{0,j}$ and the fusion parameter $f_{0,j}$ are the same across different rank $r = 1, \ldots, R$, and we denote $s_0 := \max_j s_{0,j}$ and $f_0 := \max_j f_{0,j}$. This can be easily extended to a more general case where these parameters vary with $r$. To encourage such sparse and fused components, we propose to solve the following penalized optimization problem,
\begin{eqnarray}
\min_{w_r, \bbeta_{1,r}, \ldots, \bbeta_{m+1,r}} \Big\| \cT - \sum_{r=1}^R w_r \bbeta_{1,r} \circ \ldots \circ \bbeta_{m+1,r}  \Big\|_F^2  + \; \lambda \sum_{j=1}^{m+1} \sum_{r=1}^R \left \|\Db  \bbeta_{j,r} \right\|_1, \label{eqn:opt1} \\
s.t. \;\; \|\bbeta_{j,r}\|_2 = 1~ \textrm{and}~ \|\bbeta_{j,r}\|_0 \le s_j, \; j = 1, \ldots, m+1, r = 1, \ldots, R, \nonumber 
\end{eqnarray}
for some cardinality parameters $s_1, \ldots, s_{m+1}$. Here $\|\cdot\|_0$ denotes the vector $\ell_0$ norm, i.e., the number of nonzero entries, and $\| \cdot \|_F$ denotes the tensor Frobenius norm, which is defined as $\| \cA \|_F := \sqrt{ \sum_{i_1,\ldots,i_m} \cA_{i_1,\ldots,i_m}^2}$ for a tensor $\cA \in \Rbb^{d_1\times \ldots \times d_m}$. The optimization formulation \eqref{eqn:opt1} encourages sparsity in the individual components via a direct $\ell_0$ constraint, and encourages smoothness via a fused lasso penalty. \change{We make a few remarks. First, one can easily choose which mode to impose which constraints, by modifying the penalty functions in \eqref{eqn:opt1} accordingly. See Section \ref{sec:algorithm_ours} for some specific examples in brain dynamic connectivity analysis where different constraints along different tensor modes are imposed. Second, our problem $(\ref{eqn:opt1})$ is related to a recently proposed tensor decomposition method with generalized lasso penalties in \citet{MS2016}. However, the two proposals differ in several ways. We use the $\ell_0$ truncation to achieve sparsity, whereas they used the lasso penalty. It is known that former yields an unbiased estimator, while the latter leads to a biased one in high-dimensional estimation \citep{shen2012, zhu2014}. Moreover, our rate of convergence of the parameter estimators are established under a general error tensor, whereas theirs required the error tensor to be Gaussian. Third, our method also differs from the  structured sparse principal components analysis of \citet{jenatton2010}. The latter seeks the factors that maximumly explain the variance of the data while respecting some structural constraints, and it assumes the group structure is known a priori. By contrast, our method aims to identify a low-rank representation of the tensor data, and does not require any prior structure information but learns the group structure adaptively given the data.}

\subsection{A special case: tensor Gaussian mixture model}
\label{sec:TGMM}

In model \eqref{eqn:additive-model}, no distributional assumption is imposed on the error tensor $\cE$. In this section, we show that, if one further assumes that $\cE$ is a standard Gaussian tensor, then our method reduces to a tensor version of Gaussian mixture model. 

An $m$-way tensor ${\cal X} \in \mathbb R^{d_1 \times d_2 \times \cdots \times d_m}$ is said to follow a tensor normal distribution \citep{kolda2009} with mean ${\cal M}$ and covariance matrices, $\bSigma_1, \ldots, \bSigma_m$, denoted as $\cX \sim \textrm{TN}({\cal M}; \bSigma_1, \ldots, \bSigma_m)$, if and only if $\textrm{vec}({\cal X}) \sim \textrm{N}\left(\textrm{vec}({\cal M}), \otimes_{j=1}^m  \bSigma_j\right)$, where $\textrm{vec}({\cal M}) \in \Rbb^{\prod_j d_j}$, $\textrm{vec}$ denotes the tensor vectorization, and $\otimes$ denotes the matrix Kronecker product. Following the usual definition of Gaussian mixture model, we say $\cX$ is drawn from a tensor Gaussian mixture model, if its density function is of the form, $f({\cal X})=\sum_{k=1}^K \pi_k \phi_k({\cal M}_k ;  \bSigma_{k,1}, \ldots, \bSigma_{k,m})$, where $\pi_k$ is the mixture weight, and $\phi_k({\cal M}_k ;  \bSigma_{k,1}, \ldots, \bSigma_{k,m})$ is the probability density function of a tensor Gaussian distribution $\textrm{TN}({\cal M}_k;  \bSigma_{k,1}, \ldots, \bSigma_{k,m})$. We next consider two special cases of our general tensor clustering model \eqref{eqn:additive-model}. 

First, when the error tensor $\cE$ in \eqref{eqn:additive-model} is a standard Gaussian tensor, our clustering model is equivalent to assuming the tensor-valued samples $\cX_i, i=1,\ldots,N$, follow the above tensor Gaussian mixture model. That is, 
\begin{eqnarray*}
\cX_1, \ldots, \cX_{l}  \sim  TN({\cal M}^*_1; \bI_{d_1}, \ldots, \bI_{d_m}), \;\; \ldots, \;\;
{\cal X}_{N-l}, \ldots, {\cal X}_{N}  \sim  TN({\cal M}^*_K; \bI_{d_1}, \ldots, \bI_{d_m}),
\end{eqnarray*}
where $\bI_{d}$ is a $d \times d$ identity matrix. Thus the tensor samples $\cX_1, \ldots, \cX_N$ are drawn from a tensor Gaussian mixture model with the $k$th prior probability $\pi_k = 1/K$, $k=1,\ldots,K$. 

Second, when the error tensor $\cE$ in \eqref{eqn:additive-model} is a general Gaussian tensor, our model is equivalent to assuming the samples follow
\begin{eqnarray} \label{eqn:general_data_generation}
\cX_1, \ldots, \cX_{l} \sim TN({\cal M}^*_1; \bSigma_{1,1}, \ldots, \bSigma_{1,m}), \;\; \ldots, \;\;
\cX_{N-l}, \ldots, \cX_{N}  \sim TN({\cal M}^*_K; \bSigma_{K,1}, \ldots, \bSigma_{K,m}), 
\end{eqnarray}
where $\bSigma_{k,j}$ is a general covariance matrix, $k=1,\ldots,K, j=1,\ldots,m$. 

\change{In our tensor clustering solution, we have chosen not to estimate those covariance matrices. This may lose some estimation efficiency, but it greatly simplifies both the computation and the theoretical analysis. Meanwhile, the methodology we develop can be extended to incorporate general covariances $\bSigma_{k,j}$ in a straightforward fashion, where the tensor cluster means and the covariance matrices can be estimated using a high-dimensional EM algorithm \citep{hao2018}, in which the M-step solves a penalized weighted least squares. We do not pursue this line of research in this article.}

\section{Estimation}
\label{sec:STF}

\subsection{Optimization algorithm}
\label{sec:algorithm_STD}

We first introduce some operators for achieving the sparsity and fusion structures of a given dense vector. We then present our optimization algorithm. 

The first operator is a truncation operator to obtain the sparse structure. For a vector $\bv \in \Rbb^d$ and a scaler $\tau \le d$, we define $\textrm{Truncate}(\bv, \tau)$  as
\begin{eqnarray*}
[\textrm{Truncate}(\bv, \tau)]_j = \begin{cases}  v_j & \mbox{if } j\in \textrm{supp}(\bv, \tau) \\ 0, & \mbox{otherwise }\end{cases},
\end{eqnarray*}
where $\textrm{supp}(\bv, \tau)$ refers to the set of indices of $\bv$ corresponding to its largest $\tau$ absolute values. The second is a fusion operator. For a vector $\bv \in \Rbb^d$ and a fusion parameter $\lambda >0$, the fused vector $\textrm{Fuse}(\bv, \lambda)$ is obtained via the fused lasso \citep{tibshirani2005}; i.e., 
\begin{eqnarray*}
\textrm{Fuse}(\bv, \lambda) := \arg\min_{\bu \in \mathbb R^d} \left\{ \sum_{i=1}^d (u_i - v_i)^2 + \lambda \|\Db \bu\|_1   \right\}. 
\end{eqnarray*}
An efficient ADMM-based algorithm for this fused lasso has been developed in \citet{zhu2017}. The third operator is a combination of the truncation and fusion operators. For $\bv \in \Rbb^d$ and parameters $\tau$, $\lambda$, we define $\textrm{Truncatefuse}(\bv, \tau, \lambda)$  as
\begin{eqnarray*}
\textrm{Truncatefuse}(\bv, \tau, \lambda) = \textrm{Truncate}\Big( \textrm{Fuse}(\bv, \lambda), \tau \Big).
\end{eqnarray*}
Lastly, we denote $\textrm{Norm}(\bv) = {\bv}/{\|\bv\|}$ as the normalization operator on a vector $\bv$.

\begin{algorithm}[h!]
\caption{Structured tensor factorization for optimization of \eqref{eqn:opt1}.}\label{alg:STD}
\begin{algorithmic}[1]
\STATE \textbf{Input:} tensor ${\cal T}$, rank $R$, cardinalities $(s_1, \ldots, s_{m+1})$, fusion parameters $(\lambda_1, \ldots, \lambda_{m+1})$.
\STATE For $r = 1$ \textbf{to} $R$
\STATE \hspace{0.075in} Initialize unit-norm vectors $\widehat{\bbeta}_{j,r}^{(0)}$ randomly for $j=1,\ldots,m+1$.
\STATE \hspace{0.075in} Repeat
\STATE \hspace{0.225in} For $j = 1$ \textbf{to} $m+1$
\STATE \hspace{0.35in} Update $\widehat{\bbeta}_{j,r}^{(\kappa + 1)}$ as
\begin{eqnarray}
\widetilde{\bbeta}_{j,r}^{(\kappa + 1)} &=& \change{\textrm{Norm}\Big( \cT  \times_1 \widehat{\bbeta}_{1,r}^{(\kappa+1)} \ldots \times_{j-1} \widehat{\bbeta}_{j-1,r}^{(\kappa+1)} \times_{j+1} \widehat{\bbeta}_{j+1,r}^{(\kappa)} \ldots \times_{m+1} \widehat{\bbeta}_{m+1,r}^{(\kappa)} \Big)}, \label{eqn:alg_nonsparse}\\
\check{\bbeta}_{j,r}^{(\kappa + 1)} &=& \textrm{Truncatefuse}\Big(\widetilde{\bbeta}_{j,r}^{(\kappa + 1)}, s_j, \lambda_j \Big),\label{eqn:alg_sparse}\\
\widehat{\bbeta}_{j,r}^{(\kappa + 1)} &=& \textrm{Norm}\Big(\check{\bbeta}_{j,r}^{(\kappa + 1)} \Big) \label{eqn:alg_norm}. 
\end{eqnarray}
\vspace{-0.2in}
\STATE \hspace{0.225in} End For
\STATE \hspace{0.075in} Until the termination condition is met.
\STATE \hspace{0.075in} Compute $\hat{w}_r = \cT \times_1 \widehat{\bbeta}_{1,r}^{(\kappa_{max})} \times_2 \ldots \times_m \widehat{\bbeta}_{m+1,r}^{(\kappa_{max})}$, where $\kappa_{max}$ is the terminated iteration.
\STATE \hspace{0.075in} Update the tensor 
$\cT = \cT - \hat{w}_r \widehat{\bbeta}_{1,r}^{(\kappa_{max})}  \circ \cdots \circ \widehat{\bbeta}_{m+1,r}^{(\kappa_{max})}$. 
\STATE End For
\STATE \textbf{Output:} $\hat{w}_r$ and $\widehat{\bbeta}_{j,r}^{(\kappa_{max})}$ for $j=1,\ldots,m+1$ and $r=1,\ldots,R$.
\end{algorithmic}
\end{algorithm}

We propose a structured tensor factorization procedure in Algorithm~\ref{alg:STD}. It consists of three major steps. Specifically, the step in \eqref{eqn:alg_nonsparse} essentially obtains an unconstrained tensor decomposition, and is done through the classical tensor power method \citep{anandkumar2014guaranteed}. Here, for a tensor $\cA \in \Rbb^{d_1\times \ldots \times d_m}$ and a set of vectors $\ba_j \in \Rbb^{d_j}, j = 1,\ldots,m$, the multilinear combination of the tensor entries is defined as $\cA \times_1 \ba_1 \times_2 \ldots \times_m \ba_m := \sum_{i_1\in[d_1]} \ldots \sum_{i_m\in [d_m]} a_{i_1} \ldots a_{i_m} \cA_{i_1,\ldots,i_m}  \in \Rbb$. It is then followed by our new Truncatefuse step in \eqref{eqn:alg_sparse} to generate a sparse and fused component. Finally, the step in \eqref{eqn:alg_norm} normalizes the component to ensure the unit-norm. These three steps form a full update cycle of one decomposition component.  The algorithm then updates all components in an alternating fashion, until some termination condition is satisfied. In our implementation, the algorithm terminates if the total number of iterations exceeds 20, or $\sum_{j=1}^{m+1} \|  \widehat{\bbeta}_{j,r}^{(\kappa+1)} - \widehat{\bbeta}_{j,r}^{(\kappa)} \|_2^2 \le 10^{-4}$. 

In terms of computational complexity, we note that, the complexity of operation in \eqref{eqn:alg_nonsparse} is $O(\prod_{j=1}^{m+1} d_j)$, while the complexity in \eqref{eqn:alg_sparse} consists of $O(d_j \log d_j)$ for the truncation operation and $O(d_j^3)$ for the fusion operation \citep{tibshirani2011}. \change{Therefore, the total complexity of Algorithm~\ref{alg:STD} is $O\left( R \kappa_{max} \max\{m \prod_{j=1}^{m+1} d_j, \sum_{j=1}^{m+1} d_j ^3  \} \right)$, by noting that $\sum_{j=1}^{m+1} \max \{ \prod_{j=1}^{m+1} d_j, d_j ^3  \} =  O\left( \max \{m\prod_{j=1}^{m+1} d_j, \sum_{j=1}^{m+1}d_j ^3 \} \right)$.}  It is interesting to note that, when the tensor order $m>2$ and the dimension $d_j$ along each tensor mode is of a similar value, the complexity of the sparsity and fusion operation in \eqref{eqn:alg_sparse} is to be dominated by that in \eqref{eqn:alg_nonsparse}. Consequently, our addition of the sparsity and fusion structures does not substantially increase the overall complexity of the tensor decomposition. 

Based upon the structured tensor factorization, we next summarize in Algorithm~\ref{alg:DTC} our proposed dynamic tensor clustering procedure illustrated in Figure \ref{fig:stack_tensor}. It is noted that Step 2 of Algorithm \ref{alg:DTC} utilizes the structured tensor factorization to obtain a reduced data $\widehat{\Bb}_{m+1}$, whose columns consist of all information of the original samples that are relevant to clustering. This avoids the curse of dimensionality through substantial dimension reduction. We also note that Step 2 provides a reduced data along each mode of the original tensor, and hence it is straightforward to achieve co-clustering of any single or multiple tensor modes. This is different from the classical clustering methods, where extension from clustering to bi-clustering generally requires different optimization formulations \citep[see, e.g.,][]{chi2016}.

\begin{algorithm}[t!]
\caption{Dynamic tensor clustering procedure}\label{alg:DTC}
\begin{algorithmic}[1]
\STATE \textbf{Input:} Tensor samples $\cX_1, \ldots, \cX_N \in \mathbb R^{d_1\times \cdots \times d_m}$, number of clusters $K$, cardinalities $(s_1, \ldots, s_{m+1})$, fusion parameters $(\lambda_1, \ldots, \lambda_{m+1})$, rank $R$.
\STATE \textbf{Step 1:} Stack tensor samples into a higher-order tensor ${\cal T}  \in \mathbb R^{d_1\times \cdots \times d_m\times N}$, where the $i$th slice in the last mode is $\cX_i$. 
\STATE \textbf{Step 2:} Apply Algorithm \ref{alg:STD} to ${\cal T}$ with rank $R$, cardinalities $(s_1, \ldots, s_{m+1})$, and fusion parameters $(\lambda_1, \ldots, \lambda_{m+1})$ to obtain $\widehat{\Bb}_{m+1} = (\widehat{\bbeta}_{m+1,1}^{(\kappa_{max})}, \ldots, \widehat{\bbeta}_{m+1,R}^{(\kappa_{max})}) \in \mathbb R^{N\times R}$.
\STATE \textbf{Step 3:} Apply the $K$-means clustering to $\widehat{\Bb}_{m+1}$ by treating each row as a sample.
\STATE \textbf{Output:} Cluster assignments \change{$\widehat{\cA}_1, \ldots, \widehat{\cA}_K$} from \textbf{Step 3}.
\end{algorithmic}
\end{algorithm}

\subsection{Application example: brain dynamic connectivity analysis}
\label{sec:algorithm_ours}

Our proposed dynamic tensor clustering approach applies to many different applications involving dynamic tensor. Here we consider one  specific application, brain dynamic connectivity analysis. Brain functional connectivity describes interaction and synchronization of distinct brain regions, and is characterized by a network, with nodes representing regions, and links measuring pairwise dependency between regions. This dependency is frequently quantified by Pearson correlation coefficient, and the resulting connectivity network is a region by region correlation matrix \citep{Fornito2013}. Traditionally, it is assumed that connectivity networks do not change over time when a subject is resting during the scan. However, there is growing evidence suggesting the contrary that networks are not static but dynamic over the time course of scan \citep{Hutchison2013}. To capture such dynamic changes, a common practice is to introduce sliding and overlapping time windows, compute the correlation network within each window, then apply a clustering method to identify distinct states of connectivity patterns over time \citep{AllenCalhoun2014}. 

There are several scenarios within this context, which share some similar characteristics. The first scenario is when we aim to cluster $n$ individual subjects, each represented by a $p \times p \times t$ tensor that  describes the brain connectivity pattern among $p$ brain regions over $t$ sliding and overlapping time windows. Here $n$ denotes the sample size, $p$ the number of brain regions, and $t$ the total number of moving windows. In this case, $\cT \in \Rbb^{p \times p \times t \times n}$ and $N = n$. It is natural to encourage sparsity in the first two modes of $\cT$, and thus to improve interpretability and identification of connectivities among important brain regions. Meanwhile, it is equally intuitive to encourage smoothness along the time mode of $\cT$, since the connectivity patterns among the adjacent and overlapping time windows are indeed highly correlated and similar. Toward that end, we propose the following structured tensor factorization based clustering solution, by considering the minimization problem, 
\begin{eqnarray*}
\min_{w_r, \bbeta_{1,r}, \bbeta_{2,r}, \bbeta_{3,r}  \bbeta_{4,r}} \Big\| \cT - \sum_{r=1}^R w_r \bbeta_{1,r} \circ \bbeta_{2,r}  \circ \bbeta_{3,r}  \circ \bbeta_{4,r}  \Big\|_F^2  + \lambda \sum_{r=1}^R\|\Db  \bbeta_{3,r}\|_1. \\ 
s.t. \;\; \|\bbeta_{j,r}\|_2 = 1, j = 1, \ldots, 4,  \bbeta_{1,r} = \bbeta_{2,r}, \textrm{~and~} \|\bbeta_{1,r}\|_0 \le s, r = 1, \ldots, R, 
\end{eqnarray*}
where $s$ and $\lambda$ are the cardinality and fusion parameters, respectively. This optimization problem can be solved via Algorithm \ref{alg:STD}. Given that the connectivity matrix is symmetric, i.e., $\cT$ is symmetric in its first two modes, we can easily incorporate such a symmetry constraint by setting $\widehat{\bbeta}_{2r}^{(t+1)} = \widehat{\bbeta}_{1r}^{(t+1)}$ in Algorithm \ref{alg:STD}.

Another scenario is to cluster $t$ sliding windows for a single subject, so to examine if the connectivity pattern is dynamic or not over time. Here $\cT \in \Rbb^{p \times p \times t}$ and $N = t$. In this case, we consider the following minimization problem, 
\begin{eqnarray*}
\min_{w_r, \bbeta_{1,r}, \bbeta_{2,r}, \bbeta_{3,r}} \Big\| \cT - \sum_{r=1}^R w_r \bbeta_{1,r} \circ \bbeta_{2,r}  \circ \bbeta_{3,r}  \Big\|_F^2  + \lambda \sum_{r=1}^R\|\Db  \bbeta_{3,r}\|_1. \\ 
s.t. \;\; \|\bbeta_{j,r}\|_2 = 1, j = 1, \ldots, 3,  \bbeta_{1,r} = \bbeta_{2,r}, \textrm{~and~} \|\bbeta_{1,r}\|_0 \le s, r = 1, \ldots, R. 
\end{eqnarray*}
If one is to examine the dynamic connectivity pattern for $n_g$ subjects simultaneously, then $\cT \in \Rbb^{p \times p \times n_g \times t}$. Both problems can be solved by dynamic tensor clustering in Algorithm \ref{alg:DTC}.

\subsection{Tuning}
\label{sec:tuning}

Our clustering algorithm involves a number of tuning parameters. To facilitate the computation, we propose a two-step tuning procedure. We first choose the rank $R$, the sparsity parameters $s_1,\ldots, s_{m+1}$, and the fusion parameters $\lambda_1, \ldots, \lambda_{m+1}$, by minimizing 

\begin{equation} \label{eqn:BIC}
\log\left( \frac{\|{ \cal T} -  \sum_{r\in [R]} \hat{w}_r \widehat{\bbeta}_{1,r} \circ \ldots \circ \widehat{\bbeta}_{m+1,r} \|_F^2}{\prod_{j=1}^{m+1} d_j}  \right) + \frac{\sum_{j=1}^{m+1} \log d_j}{\prod_{j=1}^{m+1} d_j} \times p_e,
\end{equation}
where $p_e$ refers to the total degrees of freedom, and is estimated by $\sum_{j=1}^{m+1} \sum_{r\in [R]}  p_e(\widehat{\bbeta}_{j,r})$, in which the degrees of freedom of an individual component $p_e(\widehat{\bbeta}_{j,r})$ is defined as the number of unique non-zero elements in $\widehat{\bbeta}_{j,r}$. For simplicity, we set $s = s_1 = \ldots = s_{m+1}$ and $\lambda = \lambda_1 = \ldots = \lambda_{m+1}$. The selection criterion \eqref{eqn:BIC} balances model fitting and model complexity, and the criterion of a similar form has been widely used in model selection \citep{wang2015}. Next, we tune the number of clusters $K$ by employing a well established gap statistic \citep{tibshirani2001}. The gap statistic selects the best $K$ as the minimal one such that $gap(k) \ge gap(k+1) - se(k+1)$, where $se(k+1)$ is the standard error corresponding to the $(k+1)$th gap statistic. In the literature, stability-based methods \citep{wang2010, wang2013} have also been proposed that can consistently select the number of clusters. We choose the gap statistic due to its computational simplicity.

\section{Theory}
\label{sec:theorem}

\subsection{Theory with $R = 1$}
\label{sec:rankone}

We first develop the theory for the rank $R = 1$ case in this section, then for the general rank case in the next section. We begin with the derivation of the rate of convergence of the general structured tensor factorization estimator from Algorithm 1. We then establish
the clustering consistency of our proposed dynamic tensor clustering in Algorithm 2. The difficulties in the technical analysis lie in the non-convexity of the tensor decomposition and the incorporation of the sparsity and fusion structures.   

Recall that we observe the tensor $\cT$ with noise as specified in \eqref{eqn:additive-model}. To quantify the noise level of the error tensor, we define the sparse spectral norm of ${\cal E}\in \mathbb R^{d_1 \times \ldots \times d_{m+1}}$ as, 
\begin{equation*}
\eta({\cal E}; s^*_1, \ldots, s^*_{m+1}) := \sup_{\substack{\| \bu_1 \| = \ldots = \|\bu_{m+1} \| = 1\\ \| \bu_1 \|_0 \le s^*_1, \ldots, \|\bu_{m+1}\|_0 \le s^*_{m+1}}} \Big|{\cal E} \times_1 \bu_1 \times_2 \ldots \times_{m+1} \bu_{m+1} \Big|,
\label{eqn:perturbation_error}
\end{equation*}
where $s_j^* \le d_j$, $j=1,\ldots,m+1$. It quantifies the perturbation error in a sparse scenario.

\begin{assumption} \label{ass:modelr1}
Consider the structure in $(\ref{eqn:CP})$. Assume the true decomposition components are sparse and smooth in that $\| \Db\bbeta_{j,r}^* \|_1 \le f_{0,j}$ and $\|\bbeta_{j,r}^*\|_0 \le s_{0j}$, for $j = 1,\ldots,m+1$. Furthermore, assume the initialization satisfies $\| \widehat{\bbeta}_{j,r}^{(0)} - \bbeta_{j,r}^*\|_2 \le \epsilon_0$, $j=1,\ldots,m$,  with $\epsilon_0 < 1$.
\end{assumption}

\noindent 
We recognize that the initialization error bound $0 \le \epsilon_0 \le 1$, since the components are normalized to have a unit norm. This condition on the initial error is very weak, by noting that we only require the initialization to be slightly better than a naive estimator whose entries are all zeros so to avoid $\epsilon_0 = 1$. As such, the proposed random initialization in our algorithm satisfies this condition with high probability. The next theorem establishes the convergence rate of the structured tensor factorization under the rank $R=1$. 

\begin{theorem}
\label{theorem:local}
Assume $R=1$ and Assumption \ref{ass:modelr1} holds. Define $M:= \max_{j} \| [\Db^{\dagger}]_j \|_2$, where $\Db^{\dagger}$ is the pseudoinverse of $\Db$ and $[\Db^{\dagger}]_j$ refers to its $j$th column. If $s_j \ge s_{0j}$, and the error tensor satisfies $\eta({\cal E}; s_1, \ldots, s_{m+1}) < w^*(1 - \epsilon_0^2)$, then $\widehat{\bbeta}_{m+1}^{(1)}$ of Algorithm \ref{alg:STD} with $\lambda_{m+1} \ge 2M \eta({\cal E}; s_1, \ldots, s_{m+1}) / [w^*(1 - \epsilon_0^2) - \eta({\cal E}; s_1, \ldots, s_{m+1})]$ satisfies, with high probability,
\begin{equation}
\| \widehat{\bbeta}_{m+1}^{(1)} - \bbeta_{m+1}^*\|^2_2 \le \left[\frac{ 2\eta({\cal E}; s_1, \ldots, s_{m+1}) }{w^*(1 - \epsilon_0^2) - \eta({\cal E}; s_1, \ldots, s_{m+1})}\right]^2 + \frac{8M f_{0,m+1}  \eta({\cal E}; s_1, \ldots, s_{m+1})}{w^*(1 - \epsilon_0^2) - \eta({\cal E}; s_1, \ldots, s_{m+1})}.
\label{eqn:bound_beta3}
\end{equation}
\end{theorem}

\noindent
With $\epsilon_0$ being a constant strictly less than 1, we see a clear tradeoff between the signal level $w^*$ and the error tensor $\cE$ according to the condition $\eta({\cal E}; s_1, \ldots, s_{m+1}) < w^*(1 - \epsilon_0^2)$. Besides, the derived upper bound in $(\ref{eqn:bound_beta3})$ reveals an interesting interaction of the initial error $\epsilon_0$, the signal level $w^*$, and the error tensor $\cE$. Apparently, the error bound can be reduced by lowering $\epsilon_0$ or $\cE$, or increasing $w^*$. For a fixed signal level $w^*$, more noisy samples, i.e., a larger $\cE$, would require a more accurate initialization in order to obtain the same error bound. Moreover, this error bound is a monotonic function of the smoothness parameter $f_{0,m+1}$. In contrast to the non-smoothed tensor factorization with $f_{0,m+1} = d_{m+1}$, our structured tensor factorization is able to greatly reduce the error bound, since $f_{0,m+1}$ is usually much smaller than $d_{m+1}$ in the dynamic setup.

Based on the general rate derived in Theorem \ref{theorem:local}, the next result shows that Algorithm \ref{alg:STD} generates a contracted estimator. It is thus guaranteed that the estimator converges to the truth as the number of iterations increases. 

\begin{corollary}
\label{thm:final}
Assume the conditions in Theorem \ref{theorem:local}, and the error tensor satisfies that 
\begin{eqnarray*}
\eta({\cal E}; s_1, \ldots, s_{m+1}) \le \min \left\{  \frac{w^* \epsilon_0 (1-\epsilon_0)}{9}, \frac{w^*\epsilon_0^2(1-\epsilon_0)}{\change{64 M f_{0,m+1} + 1}}  \right\}.
\end{eqnarray*}
If $\| \widehat{\bbeta}_{j}^{(0)} - \bbeta_j^*\|_2 \le \epsilon_0$, $j=1,\ldots,m$, then the update $\widehat{\bbeta}_{m+1}^{(1)}$ in our algorithm satisfies $\| \widehat{\bbeta}_{m+1}^{(1)} - \bbeta_{m+1}^*\|_2 \le \epsilon_0/2$ with high probability.
\end{corollary}

\noindent
Corollary \ref{thm:final} implies that the $\ell_2$ distance of the estimator to the truth from each iteration is at most half of the one from the previous iteration. Simple algebra implies that the estimator in the $\kappa$th iteration of the algorithm satisfies $\| \widehat{\bbeta}_{m+1}^{(\kappa)} - \bbeta_{m+1}^*\|_2 \le 2^{-\kappa}\epsilon_0$, which converges to zero as $\kappa$ increases. Here the assumption on $\eta({\cal E}; s_1, \ldots, s_{m+1})$ is imposed to ensure that the contraction rate of the estimator is 1/2, and it can be relaxed for a slower contraction rate.

It is also noteworthy that the above results do not require specification of the error tensor distribution. Next we derive the explicit form of the estimation error when $\cE$ is a Gaussian tensor. \change{In the following, $a_n \succ b_n$ means $b_n/a_n \rightarrow 0$, and $a_n = \tilde{O}_p(b_n)$ means $a_n, b_n$ are of the same order up to a logarithm term, i.e., $a_n = O_p(b_n (\log n)^c)$ for some constant $c>0$.}

\begin{corollary}
\label{cor:gaussian_error}
Assume the conditions in Theorem \ref{theorem:local}, and assume $\cE \in \mathbb R^{d_1 \times \ldots \times d_{m+1}}$ is a Gaussian tensor. Then we have $\eta({\cal E}; s_1, \ldots, s_{m+1}) = C\sqrt{\prod_{j=1}^{m+1} s_j \sum_{j=1}^{m+1}\log(d_j)}$ for some constant $C$. In addition, if the signal strength satisfies $w^* \succ \sqrt{\prod_{j=1}^{m+1} s_j \sum_{j=1}^{m+1}\log(d_j)}$, we have the update $\widehat{\bbeta}_{m+1}^{(1)}$ in one iteration of our algorithm satisfies
\begin{equation*}
\| \widehat{\bbeta}_{m+1}^{(1)} - \bbeta_{m+1}^*\|^2_2 = \tilde{O}_p\left( \max \left\{ \frac{\prod_{j=1}^{m+1} s_j}{w^{*2}}, \frac{f_{0,m+1} \sqrt{\prod_{j=1}^{m+1} s_j}}{w^*}   \right\}  \right). 
\label{eqn:bound_beta3_gaussian}
\end{equation*}
\end{corollary}

Next we establish the consistency of our dynamic tensor clustering Algorithm \ref{alg:DTC} under the tensor Gaussian mixture model. Consider a collection of $m$-way tensor samples, $\cX_1, \ldots, \cX_N \in \mathbb R^{d_1 \times d_2 \times \cdots \times d_m}$, from a tensor Gaussian mixture model, with $K$ rank-1 centers, $\cM_1 := \mu^*_{1}w^* \bbeta^*_{1} \circ \ldots \circ \bbeta^*_{m}, \ldots, \cM_K :=  \mu^*_{K}w^* \bbeta^*_{1} \circ \ldots \circ \bbeta^*_{m}$, and equal prior probability $\pi_k = 1/K$, for $k=1,\ldots,K$. As before, we assume an equal number of $l = N/K$ samples in each cluster. We denote the component from the last mode $\bbeta_{m+1}^*$ as,  
\begin{equation}
\bbeta_{m+1}^* = ( \underbrace{ \mu^*_{1},\ldots,  \mu^*_{1}}_{l \textrm{~samples}}, \; \ldots, \; \underbrace{\mu^*_{K}, \ldots, \mu^*_{K}}_{l \textrm{~samples}} ) \in \mathbb R^{N \times 1}.
\label{eqn:beta3_cluster}
\end{equation}
Define the true cluster assignments as $\cA_1^*:= \{1,\ldots, l\}, \ldots, \cA_K^*:= \{N-l, \ldots, N\}$. \change{Recall that the estimated cluster assignments $\widehat{\cA}_1, \ldots, \widehat{\cA}_K$ are obtained from Algorithm~\ref{alg:DTC}.} Then we show that our proposed dynamic tensor clustering estimator is consistent, in that the estimated cluster centers from our dynamic tensor clustering converge to the truth consistently, and that the estimated cluster assignments recover the true cluster structures with high probability.

\begin{theorem}
\label{thm:cluster}
Assume $R=1$ and Assumption \ref{ass:modelr1} holds. If $s_j \ge s_{0j}$, and the signal strength satisfies that $w^* \succ \sqrt{\prod_{j=1}^{m} s_j N^2 \log(\prod_{j=1}^{m} d_{j} N)/K}$, then the estimator $\widehat{\bbeta}_{m+1}$ satisfies that 

\begin{eqnarray*}
\| \widehat{\bbeta}_{m+1} - \bbeta_{m+1}^*\|_2 = O_p\left(\frac{K}{\sqrt{N}}\right).
\end{eqnarray*}
Moreover, if $\min_{i, j} |\mu^*_{i} - \mu^*_{j}| > C_1K/\sqrt{N}$ for some constant $C_1$, we have $\widehat{\cA}_k = \cA_k^*$ for any $k=1,\ldots,K$ with high probability.
\end{theorem}

\noindent
Compared to Corollary \ref{cor:gaussian_error} for the general structured tensor factorization, Theorem \ref{thm:cluster} requires a stronger condition on the signal strength $w^*$ in order to ensure the estimation error of cluster centers converges to zero at a desirable rate. Given this rate and an additional condition on the minimal gap between clusters, we are able to ensure that the estimated clusters recover the true clusters with high probability. It is also worth mentioning that our theory allows the number of clusters $K$ to diverge polynomially with the sample size $N$. \change{Besides, we allow $d_j$ and $s_j$ ($< d_j$) to diverge to infinity, as long as the signal strength condition is satisfied.}

{\color{black}

\subsection{Theory with a general rank}
\label{sec:general_rank}

We next extend the theory of structured tensor factorization and dynamic clustering to the general rank $R$ case. For this case, we need some additional assumptions on the structure of tensor factorization, the initialization, and the noise level. 

We first introduce the concept of incoherence, which is to quantify the correlation between the decomposed components.
\begin{equation}
\xi := \max_{j=1,\ldots, m+1}\max_{r \ne r^{\prime}} |\langle \bbeta_{j,r}^*, \bbeta_{j,r^{\prime}}^* \rangle|.
\label{eqn:incoherence}
\end{equation}
Denote the initialization error $\max_j\| \widehat{\bbeta}_{j,r}^{(0)} - \bbeta_{j,r}^*\|_2 \le \epsilon_0$ for some $r \in \{1,\ldots, R\}$. Denote $w_{\max} = \max_r w_r^*$ and $w_{\min} = \min_r w_r^*$. Define $g(\epsilon_0, \xi, R) := \epsilon_0^2 C_1 + 2 \epsilon_0 \xi (R-1) + \xi^2 (R-1)$. The next theorem establishes the convergence rate of the structured tensor factorization under a general rank $R$.

\begin{theorem}
\label{theorem:local_general}
Assume a general rank $R \geq 1$ and Assumption \ref{ass:modelr1} holds. Assume $\|{\cal T}^*\| \le C_1w_{\max}$, and the error tensor satisfies $\eta({\cal E}; s_1, \ldots, s_{m+1}) < w_{\min}(1 - \epsilon_0^2) - w_{\max} g(\epsilon_0, \xi, R).$ If $s_j \ge s_{0j}$, then $\widehat{\bbeta}_{m+1,r}^{(1)}$ of Algorithm \ref{alg:STD} with $\lambda_{m+1} \ge 2M [w_{\max} g(\epsilon_0, \xi, R) + \eta({\cal E}; s_1, \ldots, s_{m+1}) ]/ [w_{\min}(1 - \epsilon_0^2) - w_{\max} g(\epsilon_0, \xi, R) - \eta({\cal E}; s_1, \ldots, s_{m+1})]$ satisfies, with high probability,
\begin{eqnarray}
\| \widehat{\bbeta}_{m+1,r}^{(1)} - \bbeta_{m+1,r}^*\|^2_2 &\le& \left[\frac{ 2w_{\max} g(\epsilon_0, \xi, R) +  2\eta({\cal E}; s_1, \ldots, s_{m+1}) }{w_{\min}(1 - \epsilon_0^2) - w_{\max} g(\epsilon_0, \xi, R) - \eta({\cal E}; s_1, \ldots, s_{m+1})}\right]^2 \nonumber\\
&& + \frac{8M f_{0,m+1}w_{\max}  g(\epsilon_0, \xi, R) + 8M f_{0,m+1}  \eta({\cal E}; s_1, \ldots, s_{m+1})}{w_{\min}(1 - \epsilon_0^2) - w_{\max} g(\epsilon_0, \xi, R) - \eta({\cal E}; s_1, \ldots, s_{m+1})}.
\label{eqn:bound_beta3_general}
\end{eqnarray}
\end{theorem}

\noindent
Compared to the error bound of the $R=1$ case in Theorem \ref{theorem:local}, the error for the general rank case is slightly larger. To see this, setting $R=1$ in $(\ref{eqn:bound_beta3_general})$, we have $g(\epsilon_0, \xi, 1) =  C_1\epsilon_0^2$, and the bound in (\ref{eqn:bound_beta3_general}) is strictly larger than that in $(\ref{eqn:bound_beta3})$. This is due to an inevitable triangle inequality used in the general rank case. The derivation for the general $R$ case is more complicated, and involves a different set of techniques than the $R=1$ case.

Based on the general rate in Theorem \ref{theorem:local_general}, we then derive the local convergence which ensures that the final estimator is to converge to the true parameter at a geometric rate. For that purpose, we introduce some additional assumptions on the initialization and noise level. 

\begin{assumption}
\label{ass:initial}
For the initialization error $\max_{j,r}\| \widehat{\bbeta}_{j,r}^{(0)} - \bbeta_{j,r}^*\|_2 \le \epsilon_0$, we assume that
$$
 \epsilon_0 \le \min \left\{\frac{w_{\min}}{8C_1w_{\max}} - \frac{2\xi(R-1)}{C_1}, \;\; \frac{w_{\min}}{6w_{\max}} - \xi^2 (R-1) \right\}.
$$
\end{assumption}

\begin{assumption}
\label{ass:noise}
Assume the error tensor satisfies $\eta({\cal E}; s_1, \ldots, s_{m+1}) < w_{\min} / 6$.
\end{assumption}

\begin{assumption}
\label{ass:fuse}
Denote $f_0 = \max_j f_{0,j}$. Assume the fuse parameter satisfies
$$
f_0 \le \frac{w_{\max}  g(\epsilon_0, \xi, R)  + \eta({\cal E}; s_1, \ldots, s_{m+1})}{2Mw_{\min}(1 - \epsilon_0^2)}
$$
\end{assumption}

\noindent
Note that Assumptions \ref{ass:initial}-\ref{ass:noise} ensure that the condition $\eta({\cal E}; s_1, \ldots, s_{m+1}) < w_{\min}(1 - \epsilon_0^2) - w_{\max} g(\epsilon_0, \xi, R)$ stated in Theorem \ref{theorem:local_general} is satisfied. We also define a statistical error $\epsilon_S$ as
$$
\epsilon_S := \frac{8w_{\max}}{w_{\min}} \xi^2 (R-1) + \frac{8}{w_{\min}} \eta({\cal E}; s_1, \ldots, s_{m+1}).
$$

\begin{corollary}
\label{cor:local_general_iteration}
Assume Assumptions \ref{ass:modelr1}-\ref{ass:fuse} hold. If $s_j \ge s_{0j}$, then $\widehat{\bbeta}_{m+1,r}^{(1)}$ of Algorithm \ref{alg:STD} with $\lambda_{m+1} \ge 2M [w_{\max} g(\epsilon_0, \xi, R) + \eta({\cal E}; s_1, \ldots, s_{m+1}) ]/ [w_{\min}(1 - \epsilon_0^2) - w_{\max} g(\epsilon_0, \xi, R) - \eta({\cal E}; s_1, \ldots, s_{m+1})]$ satisfies, with high probability,
\begin{eqnarray}
\| \widehat{\bbeta}_{m+1,r}^{(1)} - \bbeta_{m+1,r}^*\|_2 \le q \epsilon_0 + \epsilon_S,
\label{eqn:bound_beta3_general_iteration}
\end{eqnarray}
where $q := 8w_{\max}[C_1\epsilon_0 + 2 \xi(R-1)]/w_{\min} \in (0,1)$. Therefore, after running $T = \Omega\{\log(\epsilon_0/\epsilon_S)\}$ iterations in Algorithm \ref{alg:STD} , we have
\begin{eqnarray}
\max_{j,r} \| \widehat{\bbeta}_{j,r}^{(T)} - \bbeta_{j,r}^*\|_2 \le O_p(\epsilon_S).
\label{eqn:bound_all_general_iteration}
\end{eqnarray}
\end{corollary}

\noindent
The error bound $(\ref{eqn:bound_beta3_general_iteration})$ ensures that the estimator in one iteration contracts at a geometric rate. It reveals an interesting interaction between the statistical error rate $\epsilon_S$ and the contraction term $q \epsilon_0$. As the iteration step $t$ increases, the computational error $q^t \epsilon_0$ decreases while the statistical error $\epsilon_S$ is fixed. After a sufficient number of iterations, the final error is to be dominated by the statistical error as shown in $(\ref{eqn:bound_all_general_iteration})$.

Next we establish the consistency of our dynamic tensor clustering Algorithm \ref{alg:DTC} under the tensor Gaussian mixture model with a general rank. Recall that our clustering analysis is conducted along the $(m+1)$th mode of the tensor, and the true parameters are defined in $(\ref{eqn:true_cluster_center})$. Analogously, we denote our dynamic tensor clustering estimator after $T$ iterations, 
as $\hat{\Bb}_{m+1} = (\hat{\bbeta}_{m+1, 1}, \ldots, \hat{\bbeta}_{m+1, R}) \in \mathbb R^{N \times R}$, and denote the $i$th row of $\hat{\Bb}_{m+1}$ as $\hat{\bmu}_i \in  \mathbb R^R$, $i=1,\ldots,N$. We quantify the clustering error via the distance between $\hat{\bmu}_i$ and $\bmu^*_i$.

\begin{theorem}
\label{thm:cluster_general}
Assume the conditions in Corollary \ref{cor:local_general_iteration} hold. Assume $w_{\max} / w_{\min} \le C_2$ for some constant $C_2>0$, the incoherence parameter satisfies $\xi^2(R-1) = O(K/\sqrt{N})$, and the minimal weight $w_{\min}$ satisfies $w_{\min} \succ \sqrt{\prod_{j=1}^{m} s_j N^2 \log(\prod_{j=1}^{m} d_{j} N)/K}$, then we have
\begin{eqnarray*}
\max_{i} \|\hat{\bmu}_i - \bmu^*_i \|_2 = O_p\left(K\sqrt{\frac{R}{N}}\right).
\end{eqnarray*}
Moreover, if $\min_{i, j} \|\bmu^*_{i} - \bmu^*_{j}\|_2 > C_3K\sqrt{R/N}$ for some constant $C_3$, we have $\widehat{\cA}_k = \cA_k^*$ for any $k=1,\ldots,K$, with high probability.
\end{theorem}

\noindent
The clustering error rate allows the true rank $R$ to increase with the sample size $N$. The clustering consistency holds as long as $K\sqrt{R/N} \rightarrow 0$. Compared to Theorem \ref{thm:cluster} in the $R=1$ case, Theorem \ref{thm:cluster_general} requires an upper bound on the incoherence parameter. A similar incoherence condition has also been imposed in \cite{anandkumar2014guaranteed} and \cite{sun2016provable} to guarantee the performance of the general-rank tensor factorization.

We also remark that, Theorem \ref{thm:cluster_general} assumes that the true rank $R$ is known. However, when the estimated rank $\widehat{R}$ exceeds the true rank $R$, those additional $\widehat{R}-R$ features can be viewed as noise. As long as their magnitudes are well controlled, it is still possible to obtain correct cluster assignments. \citet{sun2012} showed that a regularized $K$-means clustering algorithm can asymptotically achieve a consistent clustering assignment in a high-dimensional setting, where the number of features is large and many of them may contain no information about the true cluster structure. Nevertheless, their result required the features to follow a Gaussian or sub-Gaussian distribution. In our setup, the features are constructed from the low-rank tensor decomposition, and thus may not satisfy those distributional assumptions. We leave a full theoretical investigation of the rank selection consistency and the clustering consistency under an estimated rank as our future research.
}

\section{Simulations}
\label{sec:simulations}

\subsection{Setup and evaluation}

We consider two simulated experiments: clustering of two-dimensional matrix samples, and clustering of three-dimensional tensor samples. {\color{black} We assess the performance in two ways: the tensor recovery error and the clustering error. The former is defined as 
\begin{eqnarray*}
\textrm{tensor recovery error} = \frac{ \Big\| \sum_{r=1}^R\widehat{w}_r \widehat{\bbeta}_{1, r} \circ \ldots \circ \widehat{\bbeta}_{m+1, r}  - \sum_{r=1}^R w^*_r \bbeta_{1,r}^* \circ \ldots \circ \bbeta_{m+1,r}^*  \Big\|_F}{ \Big\| \sum_{r=1}^R w^*_r \bbeta_{1,r}^* \circ \ldots \circ \bbeta_{m+1,r}^*  \Big\|_F}.
\end{eqnarray*}}
The latter is defined as the estimated distance between an estimated cluster assignment $\hat{\psi}$ and the true assignment $\psi$ of the sample data ${\cal X}_1,\ldots,{\cal X}_N$, i.e., 
\begin{equation}
\textrm{clustering error} =\binom{N}{2}^{-1} \Big | \{(i,j): \ind(\widehat{\psi}({\cal X}_i)=\widehat{\psi}({\cal X}_j)) \neq \ind(\psi({\cal X}_i)=\psi({\cal X}_j)); i<j \} \Big |, \nonumber
\end{equation}
where $|A|$ is the cardinality of the set $A$. This clustering criterion has been commonly used in the clustering literature \citep{wang2010}.

We compare our proposed method with some alternative solutions. For tensor decomposition, we compare our structured tensor factorization method with two alternative decomposition solutions, the tensor truncated power method of \citet{sun2016provable}, and the generalized lasso penalized tensor decomposition method of \citet{MS2016}. For clustering, our solution is to apply $K$-means to the last component from our structured tensor factorization. Therefore, we compare with the solutions of applying $K$-means to the component from the decomposition method of \citet{sun2016provable} and \citet{MS2016}, respectively, and to the vectorized tensor data without any tensor decomposition. The solution of clustering the vectorized data is often used in brain dynamic functional connectivity analysis \citep{AllenCalhoun2014}.

\subsection{Clustering of 2D matrix data}
\label{sec:sim-2d}

We simulate the matrix observations, $\cX_i \in \mathbb R^{d_1 \times d_2}, i = 1, \ldots, N$, based on the decomposition model in $(\ref{eqn:CP})$ with a true rank $R=2$. The unnormalized components $\tilde{\bbeta}_{r,j}^*$ are generated as,
{\small
\begin{eqnarray*}
&&\tilde{\bbeta}_{1,1}^* = \tilde{\bbeta}_{2,1}^* =  (\mu, -\mu, 0.5\mu,-0.5\mu,  0,\ldots,0 ),~ \tilde{\bbeta}_{3,1}^* = ( \underbrace{\mu, \ldots, \mu}_{\lfloor N/2 \rfloor}, \; \underbrace{-\mu, \ldots, -\mu}_{\textrm{the rest}} );\\
&&\tilde{\bbeta}_{1,2}^* = \tilde{\bbeta}_{2,2}^* = (0, 0, 0, 0, \mu, -\mu, 0.5\mu,-0.5\mu, 0,\ldots,0 ), ~\tilde{\bbeta}_{3,2}^* = ( \underbrace{-\mu, \ldots, -\mu}_{\lfloor N/4 \rfloor}, \; \underbrace{\mu, \ldots, \mu}_{\lfloor N/2 \rfloor}, \; \underbrace{-\mu, \ldots, -\mu}_{\textrm{the rest}}).
\end{eqnarray*}
}

\noindent
We then obtain the true components $\bbeta_{r,j}^* = \textrm{Norm}(\tilde{\bbeta}_{r,j}^*)$, and the corresponding norm is absorbed into the tensor weight $w^*$. We set $d_1=d_2$ and vary $d_1 = \{20, 40\}$, the sample size $N = \{50, 100\}$, and the signal level $\mu = \{1, 1.2\}$. The components $\tilde{\bbeta}_{3,1}^*$ and $\tilde{\bbeta}_{3,2}^*$ determine the cluster structures of the matrix samples, resulting in four clusters, with $\cX_1, \ldots, \cX_{\lfloor N/4 \rfloor} \in \cA_1$,  $\cX_{\lfloor N/4 \rfloor+1}, \ldots, \cX_{\lfloor N/2 \rfloor} \in \cA_2$, $\cX_{\lfloor N/2 \rfloor+1}, \ldots, \cX_{\lfloor 3N/4 \rfloor} \in \cA_3$, and the rest samples belonging to $\cA_4$. 

Table \ref{tab:sim_2d} reports the average and standard error (in parentheses) of tensor recovery error, clustering error, and computational time based on 50 data replications. For tensor recovery accuracy, it is clearly seen that our method outperforms the competitors in all scenarios. When the dimension $d_1$ increases from 20 to 40, the recovery error increases, as the decomposition becomes more challenging. When the signal $\mu$ increases from 1 to $1.2$, i.e., the tensor decomposition weight $w^*$ increases, the tensor recovery error reduces dramatically. For clustering accuracy, our approach again performs the best, which demonstrates the usefulness of incorporation of both sparsity and fusion structures in tensor decomposition. \change{For  computational time, the three clustering methods based on different decompositions are comparable to each other, and are much faster than the one based on the vectorized data. It reflects the substantial computational gain of tensor factorization in this context.}

\begin{table}[h!]
\centering
\begin{small}
\caption{Clustering of 2D matrices. Reported are the average and standard error (in parentheses) of tensor recovery error, clustering error, and computational time based on 50 data replications. DTC: the proposed dynamic tensor clustering method based on structured tensor factorization (STF); TTP: the tensor truncated power method of \citet{sun2016provable}; GLTD: the generalized lasso penalized tensor decomposition method of \citet{MS2016}; vectorized: $K$-means clustering based on the vectorized data.}
\label{tab:sim_2d}
\vskip 1 em
\begin{tabular}{ccc|cccc} \hline
 & & & \multicolumn{4}{c}{Tensor recovery error} \\ \cline{4-7}
  $d_1 = d_2$ & $N$ & $\mu$ & STF & TTP & GLTD \\ \hline
20 & 50 & 1 & \textbf{0.833} (0.012) & 0.879 (0.017) & 0.938 (0.018) \\ 
 &  & 1.2 & \textbf{0.305} (0.016) & 0.427 (0.031) & 0.474 (0.032) \\ 
 & 100 & 1 & \textbf{0.786} (0.008) & 0.826 (0.015) & 0.886 (0.019) \\ 
 &  & 1.2 & \textbf{0.284} (0.016) & 0.416 (0.032) & 0.505 (0.036) \\ 
40 &   50 & 1 & \textbf{0.883} (0.016) & 0.999 (0.016) & 1.040 (0.011) \\ 
 &   & 1.2 & \textbf{0.498} (0.032) & 0.696 (0.039) & 0.802 (0.032) \\ 
 &  100 & 1 & \textbf{0.867} (0.017) & 0.963 (0.018) & 1.009 (0.014) \\ 
 &   & 1.2 & \textbf{0.371} (0.027) & 0.552 (0.037) & 0.694 (0.036) \\ 
\hline
& & & \multicolumn{4}{c}{Clustering error} \\ \cline{4-7}
 $d_1 = d_2$ &$N$ & $\mu$ & DTC & TTP & GLTD & vectorized \\ 
  \hline
 20 &  50 & 1 & \textbf{0.291} (0.010) & 0.332 (0.015) & 0.382 (0.017) & 0.306 (0.006) \\ 
 &   & 1.2 & \textbf{0.015} (0.009) & 0.076 (0.017) & 0.093 (0.018) & 0.255 (0.000) \\ 
 &  100 & 1 & \textbf{0.270} (0.007) & 0.304 (0.013) & 0.351 (0.017) & 0.282 (0.002) \\ 
 &   & 1.2 & \textbf{0.015} (0.009) & 0.076 (0.017) & 0.121 (0.019) & 0.253 (0.000) \\ 
 40 &  50 & 1 & \textbf{0.337} (0.015) & 0.440 (0.015) & 0.466 (0.013) & 0.429 (0.007) \\ 
 &   & 1.2 & \textbf{0.118} (0.019) & 0.242 (0.027) & 0.292 (0.025) & 0.262 (0.002) \\ 
 &  100 & 1 & \textbf{0.336} (0.016) & 0.415 (0.016) & 0.446 (0.014) & 0.376 (0.006) \\ 
 &   & 1.2 & \textbf{0.061} (0.015) & 0.151 (0.022) & 0.227 (0.024) & 0.253 (0.000) \\ 
\hline
& & & \multicolumn{4}{c}{Computational time (seconds)} \\ \cline{4-7}
 $d_1 = d_2$ &$N$ & $\mu$ & DTC & TTP & GLTD & vectorized \\ 
  \hline
20 & 50 & 1 & 1.612 (0.325) & \textbf{1.097} (0.064) & 1.662 (0.067) & 43.85 (0.030) \\ 
   &  & 1.2 & 2.581 (0.456) & \textbf{1.598} (0.089) & 2.669 (0.095) & 43.86 (0.022) \\ 
   & 100 & 1 & 1.971 (0.370) & \textbf{1.407} (0.075) & 1.995 (0.077) & 100.2 (0.059) \\ 
   &  & 1.2 & 3.811 (0.882) & \textbf{2.717} (0.173) & 3.962 (0.184) & 123.4 (0.113) \\ 
  40 & 50 & 1 & 2.956 (0.762) & \textbf{1.446} (0.150) & 3.251 (0.155) & 290.7 (0.502) \\ 
   &  & 1.2 & 4.432 (0.999) & \textbf{1.890} (0.203) & 4.745 (0.276) & 290.2 (0.518) \\ 
   & 100 & 1 & 3.410 (0.956) & \textbf{1.954} (0.188) & 3.696 (0.195) & 569.4 (2.979) \\ 
   & & 1.2 & 5.920 (1.509) & \textbf{3.155} (0.307) & 6.456 (0.368) & 572.5 (2.047) \\ 
\hline
\end{tabular}
\end{small}
\end{table}

\subsection{Clustering of 3D tensor data}
\label{sec:sim-3d}

We simulate the three-dimensional tensor samples, $\cX_i \in \mathbb R^{d_1 \times d_2 \times d_3}, i = 1, \ldots, N$. We vary the sample size $N = \{50, 100\}$, the signal level $\mu = \{0.6, 0.8\}$, and set $d_1 = d_2 = d_3 = 20$. We did not consider $d_1=40$, since the computation is too expensive for the alternative solution that applies $K$-means to the vectorized data. The unnormalized components are generated as, 
{\small
\begin{eqnarray*}
&&\tilde{\bbeta}_{1,1}^* = \tilde{\bbeta}_{2,1}^* = \tilde{\bbeta}_{3,1}^* =  (\underbrace{\mu, \ldots, \mu}_{5}, \; \underbrace{-\mu, \ldots, -\mu}_{5}, \; \underbrace{0, \ldots, 0}_{10} ), \; \tilde{\bbeta}_{4,1}^* = ( \underbrace{\mu, \ldots, \mu}_{\lfloor N/2 \rfloor}, \; \underbrace{-\mu, \ldots, -\mu}_{\textrm{the rest}} );\\
&&\tilde{\bbeta}_{1,2}^* = \tilde{\bbeta}_{2,2}^* = \tilde{\bbeta}_{3,2}^* =  (\underbrace{0, \ldots, 0}_{10}, \; \underbrace{\mu, \ldots, \mu}_{5}, \; \underbrace{-\mu, \ldots, -\mu}_{5} ), \; \tilde{\bbeta}_{4,2}^* = ( \underbrace{-\mu, \ldots, -\mu}_{\lfloor N/4 \rfloor}, \; \underbrace{\mu, \ldots, \mu}_{\lfloor N/2 \rfloor}, \; \underbrace{-\mu, \ldots, -\mu}_{\textrm{the rest}} ).
\end{eqnarray*}
}

\begin{table}[h!]
\centering
\begin{small}
\caption{Clustering of 3D tensors. The methods under comparison are the same as described in Table \ref{tab:sim_2d}.}
\label{tab:sim_3d}
\vskip 1 em
\begin{tabular}{ccc|cccc} \hline
& & & \multicolumn{4}{c}{Tensor recovery error} \\ \cline{4-7}
 $d_1 = d_2 = d_3$ & $N$ & $\mu$ & STF & TTP & GLTD \\ \hline
20 & 50 & 0.6 & \textbf{0.557} (0.054) & 1.073 (0.004) & 0.688 (0.060)\\ 
 &   & 0.8 & \textbf{0.083} (0.001) & 0.253 (0.059) & 0.214 (0.058) \\ 
  & 100 & 0.6 & \textbf{0.415} (0.055) & 0.821 (0.044) & 0.493 (0.056) \\ 
 &  & 0.8 & \textbf{0.081} (0.001) & 0.465 (0.070) & 0.116 (0.031) \\ \hline
 & & & \multicolumn{4}{c}{Clustering error} \\ \cline{4-7}
  $d_1 = d_2 = d_3$ &$N$ & $\mu$ & DTC & TTP & GLTD & vectorized \\ 
\hline
20 &  50 & 0.6 & \textbf{0.154} (0.029) & 0.443 (0.024) & 0.302 (0.034) & 0.397 (0.014) \\ 
 &    & 0.8 & \textbf{0.000} (0.000) & 0.049 (0.023) & 0.051 (0.023) & 0.255 (0.000) \\ 
  & 100 & 0.6 & \textbf{0.076} (0.027) & 0.332 (0.032) & 0.176 (0.037) & 0.314 (0.006) \\ 
 &  & 0.8 & \textbf{0.000} (0.000) & 0.148 (0.028) & 0.013 (0.013) & 0.253 (0.000) \\ 
\hline
 & & & \multicolumn{4}{c}{Computational time (seconds)} \\ \cline{4-7}
  $d_1 = d_2 = d_3$ &$N$ & $\mu$ & DTC & TTP & GLTD & vectorized \\ 
\hline
20 &  50 & 0.6 & 5.847 (0.436) & \textbf{4.642} (0.323) & 6.048 (0.457) & 1746 (2.897) \\ 
 &    & 0.8 & 5.974 (0.129) & \textbf{5.347} (0.145) & 6.623 (0.199) & 1716 (1.172) \\ 
  & 100 & 0.6 & 11.04 (0.688) & \textbf{9.923} (0.592) & 11.41 (0.706) & 3849 (36.93) \\ 
 &  & 0.8 & 9.774 (0.367) & \textbf{9.516} (0.304) & 10.23 (0.349) & 3696 (35.31) \\ 
\hline
\end{tabular}
\end{small}
\end{table}

\begin{figure}[h!]
\centering
\includegraphics[scale=0.4]{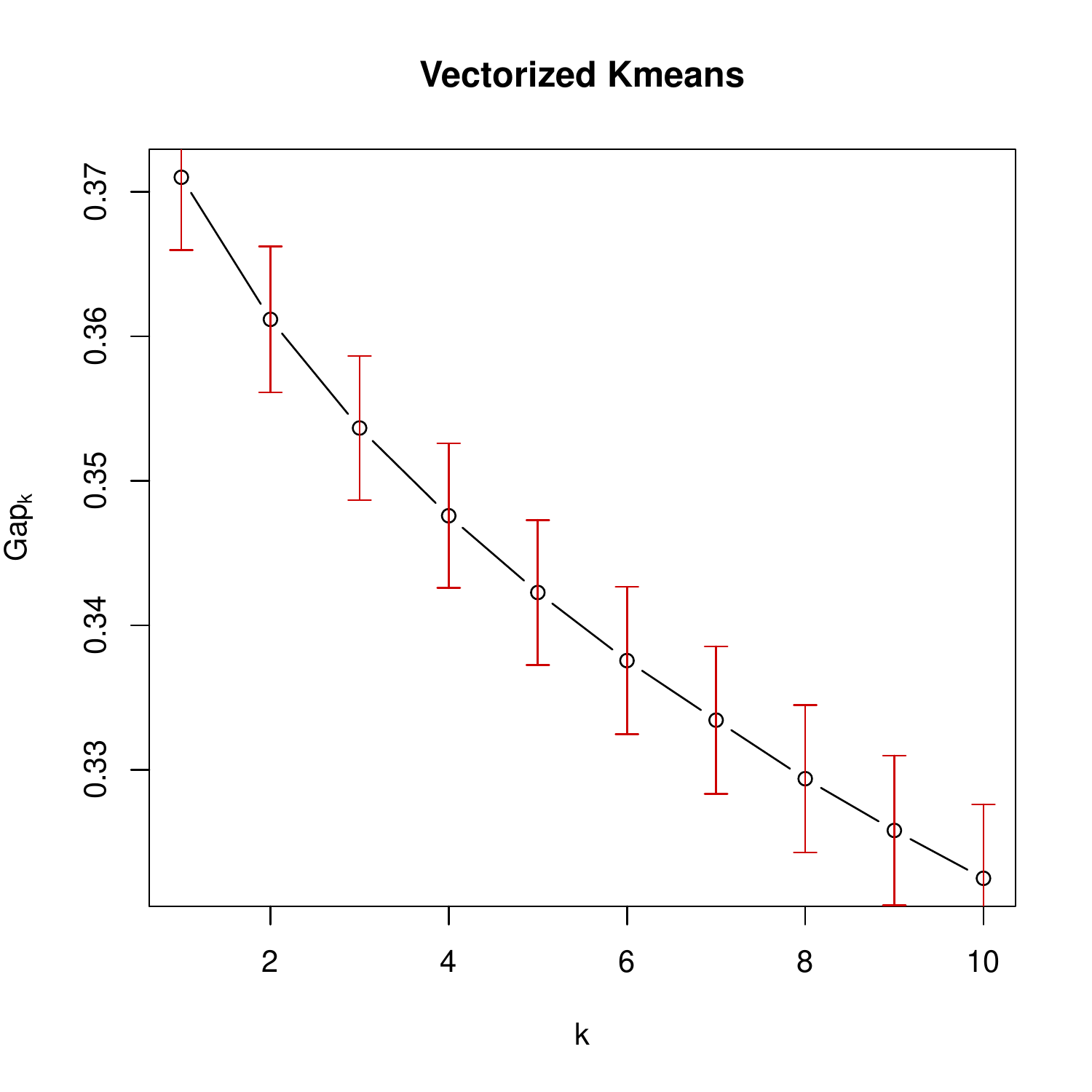}~~
\includegraphics[scale=0.4]{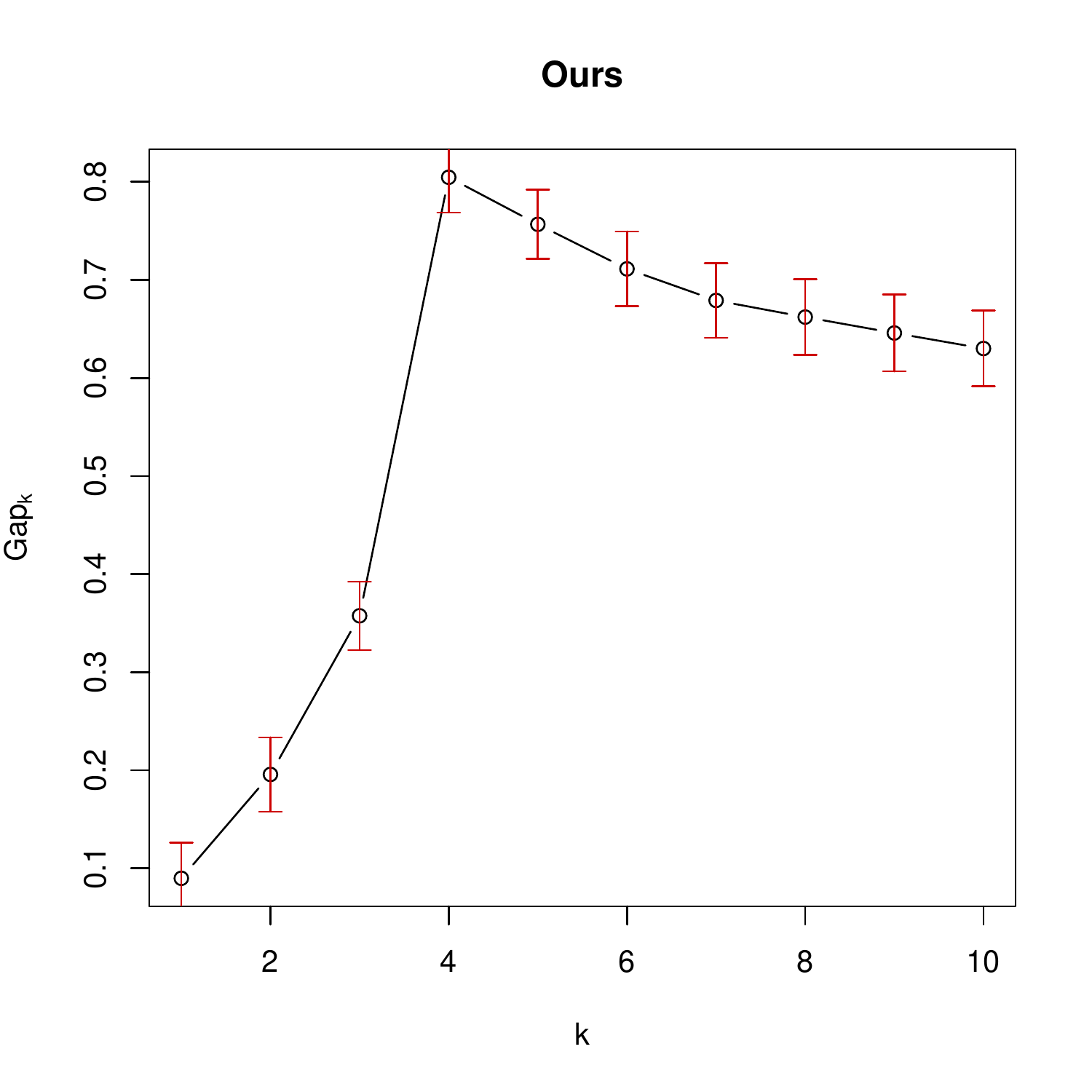}
\caption{Gap statistics of clustering 3D tensor samples based on a single data replication. Left panel shows the gap statistic from applying $K$-means to the vectorized tensor data, and the right panel shows that after structured tensor factorization. The true number of clusters is 4 in this example.}
\label{fig:gap_3d}
\end{figure}

Table \ref{tab:sim_3d} reports the average and standard error (in parentheses) of tensor recovery error, clustering error, and computational time based on 50 data replications. Again we observe similar qualitative patterns in 3D tensor clustering as in 2D matrix clustering, except that the advantage of of our method is even more prominent. It is also noted that the clustering error of applying $K$-means to the vectorized data without any decomposition is at least 0.253 even when the signal strength increases. This is due to the fact that the method always mis-estimates the number of clusters $K$. By contrast, our method could estimate $K$ correctly. Figure \ref{fig:gap_3d} illustrates the gap statistic of both methods in one data replication with $\mu=0.6$, where the true value $K = 4$ in this example. Moreover, the tuning time for the method with the vectorized data is very long due to expensive computation of gap statistic in this ultrahigh dimensional setting. This example suggests that our method of clustering after structured tensor factorization has clear advantages in not only computation but also the tuning and the subsequent clustering accuracy compared to the simple alternative solution that directly applies clustering to the vectorized data.

\change{In the interest of space, additional simulations with different correlation structures, large ranks, and unequal cluster sizes are reported in Section \ref{sec:additional_experiments} of the Supplementary Materials. }

\section{Real data analysis}
\label{sec:realdata}

We illustrate our dynamic tensor clustering method through a brain dynamic connectivity analysis based on resting-state functional magnetic resonance imaging (fMRI). Meanwhile we emphasize that our proposed method can be equally applied to many other dynamic tensor applications as well. The data is from the Autism Brain Imaging Data Exchange (ABIDE), a study of autism spectrum disorder (ASD) \citep{DiMartino2014}. ASD is an increasingly prevalent neurodevelopmental disorder, characterized by symptoms such as social difficulties, communication deficits, stereotyped behaviors and cognitive delays \citep{rudie2013}. The data were obtained from multiple imaging sites. We have chosen to focus on the fMRI data from the \change{University of Utah School of Medicine (USM)} site only, since the sample size at USM is relatively large; meanwhile it is not too large so to ensure the computation of applying $K$-means clustering to the vectorized data is feasible. See more discussion on the computational aspect in our first task and Table \ref{tab:USM_error}. The data consists of resting-state fMRI of 57 subjects, of whom 22 have ASD, and 35 are normal controls. The fMRI data has been preprocessed, and is summarized as a $116 \times 236$ spatial-temporal matrix for each subject. It  corresponds to 116 brain regions-of-interest from the Anatomical Automatic Labeling (AAL) atlas, and each time series is of length 236. We consider two specific tasks that investigate brain dynamic connectivity patterns using the sliding window approach. 

\change{Specifically, in the first task, we aim to cluster the subjects based on their dynamic brain connectivity patterns, then  compare our estimated clusters with the subject's diagnosis status, which is treated as the true cluster membership in this analysis.} The targeting tensor is $\cT \in \Rbb^{116 \times 116 \times t \times n}$, which stacks a sequence of correlation matrices of dimension $116 \times 116$ corresponding to 116 brain regions over $t$ sliding time windows, and $n$ is the total number of subjects and equals 57 in this example. There are two parameters affecting the moving windows, the width of the window and the step size of each movement. We combined these two into a single parameter, the number of moving windows $t$, by fixing the width of moving window at 20. We varied the value of $t$ among $\{1, 30, 50, 80\}$ to examine its effect on the subsequent clustering performance. \change{When $t=1$, it reduces to the usual static connectivity analysis where the correlation matrix is computed based on the entire spatial-temporal matrix.} \change{We clustered along the last mode of $\cT$, i.e., the mode of individual subjects. Meanwhile, we imposed fusion smoothness along the time mode, since by construction, the connectivity patterns of the adjacent and overlapping sliding windows are very similar. We did not impose any sparsity constraint, since our focus in this task is on the overall dynamic connectivity behavior rather than individual connections. To compare with the diagnosis status, we fixed the number of clusters at $K=2$. We report the clustering error of our method in Table \ref{tab:USM_error}, along with the alternative methods in Section \ref{sec:simulations}.} We make a few observations. First, the clustering error based on dynamic connectivity ($t > 1$) is consistently better than the one based on static connectivity ($t = 1$). This suggests that the underlying connectivity pattern is likely dynamic rather than static for this data. Second, the clustering accuracy of our method is considerably better than that of applying $K$-means to the vectorized data. Moreover, our method is computationally much more efficient. \change{Actually, the method that directly applied $K$-means to the vectorized data ran out of memory on a personal laptop computer in the case of $t=80$ since the space complexity of Lloyd's algorithm for $K$-means is $O((K+n)d)$ \citep{hartigan1979}, where the number of clusters $K=2$, the sample size $n=57$, and the dimension of the vectorized tensor $d=116 \times 116 \times 80 \sim 10^6$. We later conducted analysis on a computer cluster.} Third, the number of moving windows does indeed affect the clustering accuracy, as one would naturally expect. In practice, we recommend experimenting with multiple values of $t$. Furthermore, we can treat $t$ as another tuning parameter. \change{To mitigate temporal inconsistency, we have also conducted the analysis in the frequency domain, following \citet{ahn2015, calhoun2003}. We report the results in Section \ref{sec:additional_experiments} of the Supplementary Materials.}

\begin{table}[t]
\centering
\begin{small}
\caption{Clustering of the ABIDE data along the \emph{subject mode}. Reported are clustering errors with different sliding windows. The methods under comparison are the same as described in Table \ref{tab:sim_2d}.}
\label{tab:USM_error}
\vskip 1 em
\begin{tabular}{c|cccc} \hline
Windows & DTC & TTP & GLTD & vectorized \\
\hline
1 & 26/57 = 0.456 & 26/57 = 0.456 &   27/57 = 0.474 & 27/57 = 0.474 \\
30 & 22/57 = 0.386 & 25/57 = 0.439 &   25/57 = 0.439 &   27/57 = 0.474 \\
50 & 18/57 = 0.316 & 21/57 = 0.368 & 21/57 = 0.368 &  28/57 = 0.491 \\
80 & \textbf{15/57 = 0.263} & \textbf{15/57 = 0.263} & \textbf{15/57 = 0.263} & \change{21/57 = 0.368}  \\
\hline
\end{tabular}
\end{small}
\end{table}

\begin{figure}[h!]
\centering
\vskip -0.5em
\includegraphics[scale=0.55]{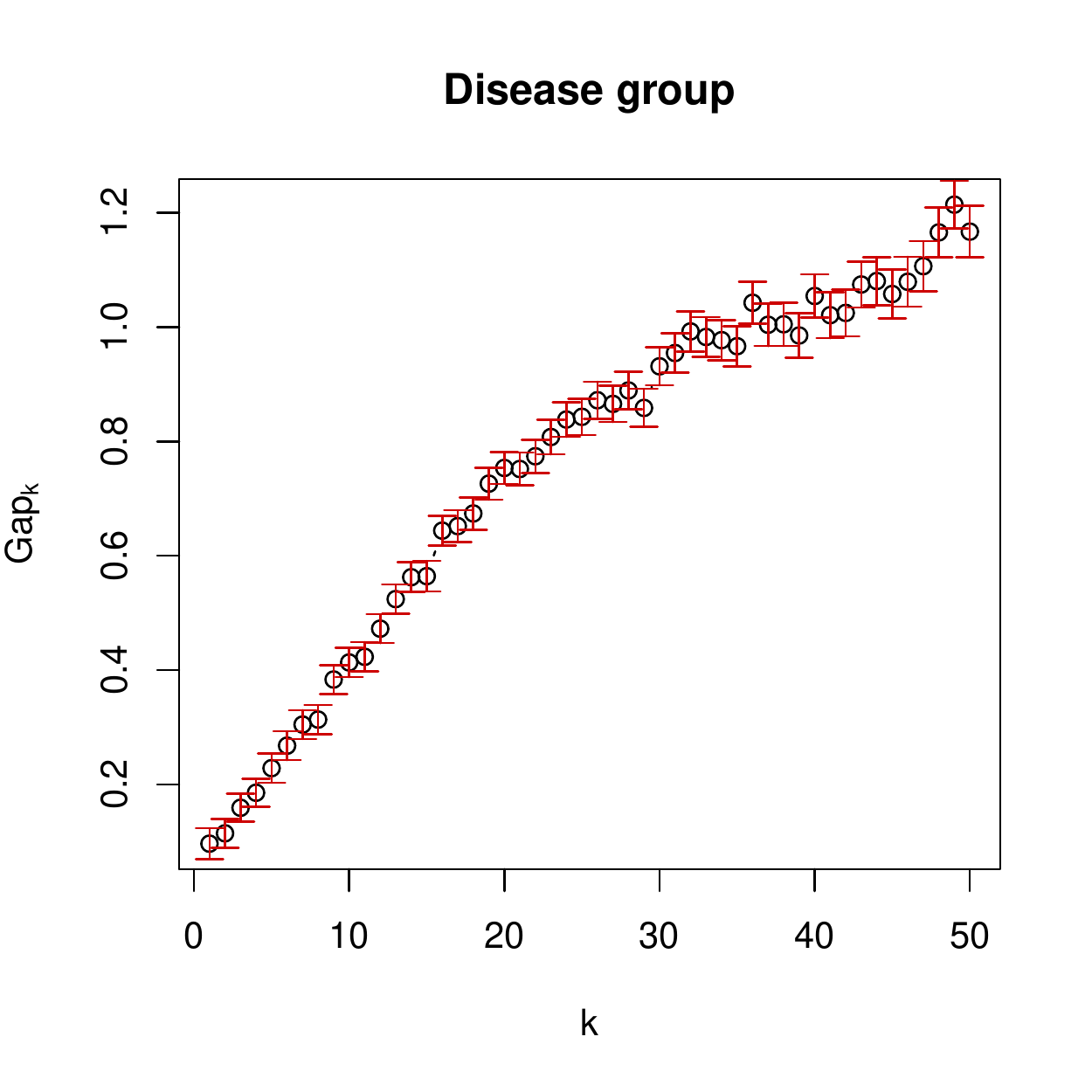}
\includegraphics[scale=0.55]{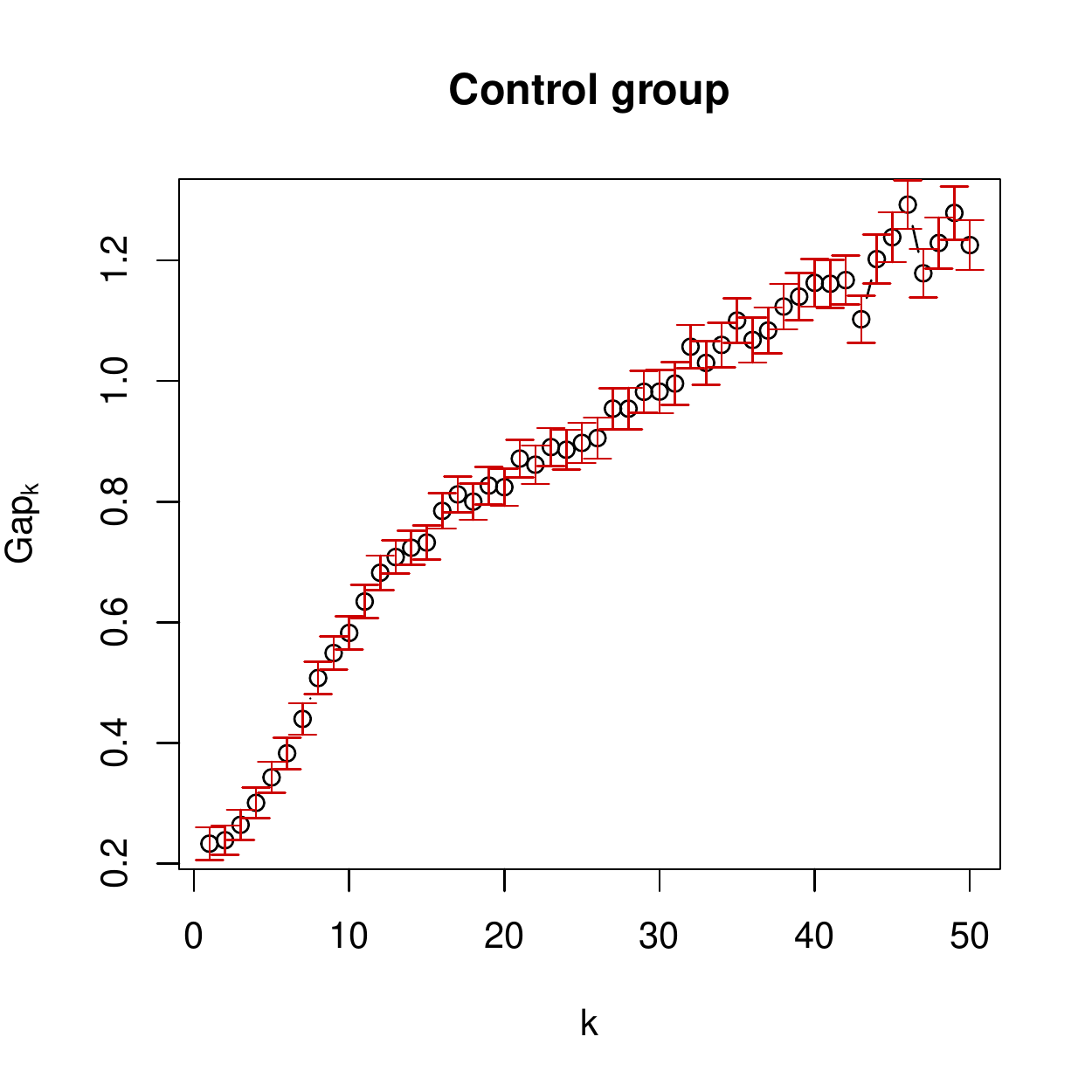}
\vskip -0.5em
\caption{Clustering of the ABIDE data along the \emph{time mode}. The gap statistic as a function of the number of clusters for the ASD group and the control group. It chose $K = 7$ for the ASD group, and $K = 12$ for the control group.}
\label{fig:USM_gap}
\end{figure}

In the second task, we aim to uncover the dynamic behavior of the brain connectivity of each diagnosis group and compare between the ASD group with the normal control.  The targeting tensor is $\cT \in \Rbb^{116 \times 116 \times n_g \times t}$, where $n_g$ denotes the number of subjects in each diagnosis group; in our example, $n_1 = 22$ and $n_0 = 35$. In light of the results from the first task, we fixed the number of moving time windows at $t = 80$. \change{We clustered along the last mode of $\cT$, i.e., the time mode, to examine the dynamic behavior, if any, of the functional connectivity patterns. We imposed the sparsity constraint on the first two modes to improve interpretation and to identify important connections among brain regions. We imposed the fusion constraint on the time mode to capture smoothness of the connectivities along the adjacent sliding windows.} We carried out two clustering analysis, one for each diagnostic group. We employed the selection criterion \eqref{eqn:BIC} to select the rank, the sparsity parameter, and the fusion parameter, and then tuned the number of clusters $K$ using the gap statistic. For both the ASD and the normal control group, the selected rank is 5. The selected sparsity parameter is 0.9, and the fusion parameter is 0.5, suggesting that in the estimated decomposition components, there are $10\%$ zero entries, and about $40$ unique values along the time mode. Figure \ref{fig:USM_gap} further shows the gap statistic as a function of the number of clusters for both diagnostic groups. Accordingly, it selects $K = 7$ for the ASD group, and $K = 12$ for the normal control. Moreover, we have observed that, for the ASD group, the clustering membership changed $10$ times along the sliding windows, whereas it changed $14$ times for the control. This on one hand suggests that both groups of subjects exhibit dynamic connectivity changes over time. More importantly, the change of the state of connectivity is less frequent for the ASD group than the control. This finding agrees with the literature in that the ASD subjects are usually found less active in brain connectivity patterns \citep{solomon2009, gotts2012, rudie2013}.

\baselineskip=13pt
\bibliographystyle{ims}
\bibliography{dynamic}

\newpage
\baselineskip=18pt

\setcounter{page}{1}
\setcounter{section}{0}
\setcounter{equation}{0}
\pagenumbering{arabic}
\renewcommand{\thesubsection}{S.\arabic{subsection}}
\renewcommand\theequation{S\arabic{equation}}
\renewcommand\thetable{S\arabic{table}}
\renewcommand\thefigure{S\arabic{figure}}

\begin{center}
\Large{\bf Supplementary Materials for \\
``Dynamic Tensor Clustering"}
\end{center}
\smallskip

\noindent
This supplementary note collects auxiliary lemmas, detailed proofs for the theorems and corollaries in Section \ref{sec:theorem}, and additional numerical analyses.

\subsection{Auxiliary lemmas} 

Lemma \ref{lemma:local} provides the error bound of a fused lasso estimator in trend filtering. 
\begin{lemma}
\label{lemma:local}
Consider the model $\yb = \bbeta^* + \bepsilon$ with true parameter $\bbeta^* \in \mathbb R^{d}$ and any noise $\bepsilon$. Denote the fused lasso estimator as $\widehat{\bbeta} := \arg\min_{\bbeta} \frac{1}{2} \| \yb - \bbeta\|_2^2 + \lambda \| \Db \bbeta \|_1$. Denote $M:= \max_{j} \| [\Db^{\dagger}]_j \|_2$. If $\lambda \ge M \| \bepsilon \|_{\infty}$, then we have
\begin{equation}
\frac{1}{d}\| \widehat{\bbeta} - \bbeta^*\|_2^2 \le \frac{\| \bepsilon \|_{\infty}^2}{d} + \frac{4\lambda \| \Db \bbeta^* \|_1}{d}
\label{eqn:lemma_bound_beta}
\end{equation}
\end{lemma}

\noindent
Its proof follows from similar arguments to the proof of Theorem 3 in \citet{wang2016}, and hence is omitted. The difference is that here we consider a general error $\bepsilon$, while they considered a Gaussian error. When $\bepsilon$ is Gaussian, Lemma \ref{lemma:local} reduces to their Theorem 3, by applying the standard Gaussian tail inequality to bound $\| \bepsilon \|_{\infty}$.  

Lemma \ref{lemma:concentration} provides a concentration of Lipschitz functions of Gaussian random variables \citep{massart2003}. 
\begin{lemma}
\label{lemma:concentration}
\citep[Theorem 3.4]{massart2003}
Let $\vb\in \mathbb R^d$ be a Gaussian random variable such that $\vb \sim N(0, \Ib_d)$. Assuming $g(\vb) \in \mathbb R$ to be a Lipschitz function such that $|g(\vb_1) - g(\vb_2)| \le L\|\vb_1 - \vb_2\|_2$ for any $\vb_1,\vb_2\in \mathbb R^d$, then we have, for each $t>0$,
$$
\mathbb P\left[ |g(\vb) - \mathbb E[g(\vb)]|  \ge t \right] \le 2 \exp\left(-\frac{t^2}{2L^2} \right).
$$ 
\end{lemma}

Lemma \ref{lemma:gaussian_width} provides an upper bound of the Gaussian width of the unit ball for the sparsity regularizer \citep{raskutti2017convex}.
\begin{lemma}
\label{lemma:gaussian_width}
For a tensor ${\cal T}\in \mathbb R^{d_1 \times d_2 \times d_3}$, denote its regularizer $R({\cal T}) = \sum_{j_1}\sum_{j_2}\sum_{j_3 } |{\cal T}_{j_1,j_2,j_3}|$. Define the unit ball of this regularizer as $B_{R}(1) := \{{\cal T}\in \mathbb R^{d_1 \times d_2 \times d_3}| R({\cal T}) \le 1 \}$. For a tensor ${\cal G} \in \mathbb R^{d_1 \times d_2 \times d_3}$ whose entries are independent standard normal random variables, we have
$$
\mathbb E\left[ \sup_{{\cal T} \in B_{R}(1)  } \langle {\cal T}, {\cal G} \rangle  \right] \le  c\sqrt{ \log(d_1d_2d_3)},
$$
for some bounded constant $c>0$.
\end{lemma}
Lemma \ref{lemma:L1normbound} links the hard thresholding sparsity and the $L_1$-penalized sparsity.
\begin{lemma}
\label{lemma:L1normbound}
For any vectors $\ub \in \mathbb R^{d_1}, \vb  \in \mathbb R^{d_2}, \wb  \in \mathbb R^{d_3}$ satisfying $\| \ub \|_2 = \|\vb\|_2 = \|\wb\|_2 = 1, \| \ub \|_0 \le s_1, \|\vb\|_0 \le s_2$, and $\|\wb\|_0 \le s_3$, denoting $\cA:=  \ub \circ \vb \circ \wb$, we have
$$ 
\|\cA\|_1:=\sum_{j_1}\sum_{j_2}\sum_{j_3 } |{\cal A}_{j_1j_2j_3}| \le \sqrt{s_1s_2s_3}.
$$ 
\end{lemma}

\textit{Proof}: According to the Cauchy-Schwarz inequality, we have $\|\ub\|_1 \le \sqrt{s_1} \|\ub\|_2 = \sqrt{s_1}$, and $\|\vb\|_1 \le \sqrt{s_2}$, $\|\wb\|_1 \le \sqrt{s_3}$. Therefore, $\|\cA\|_1 = \|\ub \circ \vb \circ \wb\|_1 \le \|\ub\|_1 \cdot \|\vb\|_1 \cdot \|\wb\|_1 \le \sqrt{s_1s_2s_3}$. \eop

Lemma \ref{lemma:pre2} connects the non-sparse entries in a tensor to the non-sparse entries in the low-rank factorization components. 
\begin{lemma}
\label{lemma:pre2}
\citep[Lemma S.6.2]{sun2016provable}
For any tensor ${\cal T} \in \mathbb{R}^{d_1 \times d_2 \times d_3}$ and an index set $F = F_1 \circ F_2 \circ F_3$ with $F_i \subseteq \{1,\ldots,d\}$, if ${\cal T} = \sum_{i\in [R]} w_i \ab_i \circ \bbb_i \circ \cbb_i$, we have
$$
{\cal T}_F = \sum_{i\in [R]} w_i \textrm{Truncate}(\ab_i, {F_1}) \circ \textrm{Truncate}(\bbb_i, {F_2}) \circ \textrm{Truncate}(\cbb_i, {F_3}).
$$
\end{lemma}

\subsection{Proof of Theorem \ref{theorem:local}} 

Throughout the supplementary materials, we prove the case with $m=2$. The extension to a general $m$ follows immediately. Also, for notational simplicity, we denote the one-step estimator $\widehat{\bbeta}_{3}^{(1)}$ as $\widehat{\bbeta}_{3}$. Note that in the update of $\widehat{\bbeta}_{3}$ in our algorithm, we first obtain an unconstrained estimator $\tilde{\bbeta}_{3}$ in $\eqref{eqn:alg_nonsparse}$, then apply the truncation and fusion operator to $\tilde{\bbeta}_{3}$. Therefore, our estimator $\widehat{\bbeta}_{3}$ is also a solution to the following problem,
$$
\widehat{\bbeta}_3 := \arg\min_{\bbeta \in \cS(d_3, s_3)} \frac{1}{2} \| \tilde{\bbeta}_{3} - \bbeta\|_2^2 + \lambda_3 \| \Db \bbeta \|_1.
$$
The underlying true model corresponding to the above problem is assumed to be
\begin{equation*}
\tilde{\bbeta}_{3} = \bbeta_3^* + \bepsilon,
\label{eqn:beta3_model}
\end{equation*}
where $\bepsilon$ is some error term. Note that the distribution of $\bepsilon$ is unknown. Therefore, in order to derive the rate of $\| \widehat{\bbeta}_{3} - \bbeta_3^*\|_2$, we utilize the result of the fused lasso problem with a general error as shown in Lemma \ref{lemma:local}.

According to the fusion assumption, we have $\|\Db \bbeta_3^* \|_1 \le f_{03}$. To derive the explicit form of the estimation error, Lemma \ref{lemma:local} suggests to calculate $\|\bepsilon\|_{\infty}$. By the definition of $\tilde{\bbeta}_{3}$, we have
$$
\bepsilon = \textrm{Norm}\big({\cal T} \times_1 \widehat{\bbeta}_{1} \times_2  \widehat{\bbeta}_{2}   \big) - \bbeta_3^*.
$$
Denote $F_1 := \textrm{supp}(\bbeta_1^*) \cup \textrm{supp}(\widehat{\bbeta}_{1})$, $F_2 := \textrm{supp}(\bbeta_2^*) \cup \textrm{supp}(\widehat{\bbeta}_2)$, and $F_3 := \textrm{supp}(\bbeta_3^*) \cup \textrm{supp}(\widehat{\bbeta}_3)$, where $\textrm{supp}(\bv)$ refers to the set of indices in $\bv$ that are nonzero. Let $F := F_1 \circ F_2 \circ F_3$. Consider the following update,
\begin{equation*}
\tilde{\bbeta}_3^{'} = \textrm{Norm}\big({\cal T}_F \times_1 \widehat{\bbeta}_{1} \times_2  \widehat{\bbeta}_{2}   \big),
\label{eqn:update_beta3_reformulate}
\end{equation*}
where ${\cal T}_F$ denotes the restriction of tensor ${\cal T}$ on the three modes indexed by $F_1$, $F_2$ and $F_3$. Note that replacing $\tilde{\bbeta}_3$ with $\tilde{\bbeta}_3^{'}$ in our algorithm does not affect the iteration of $ \widehat{\bbeta}_{3}$ due to the sparsity restriction of ${\cal T}_F$ and the scaling-invariant truncation operation. Therefore, in the sequel, we assume that $\tilde{\bbeta}_3$ is replaced by $\tilde{\bbeta}_3^{'}$.

Plugging the underlying model ${\cal T} = {\cal T}^* + {\cal E}$ with the rank-1 true tensor $\cT^* =   w^* \bbeta^*_{1} \circ \bbeta^*_{2} \circ \bbeta^*_{3}$ into the above expression, we have,
\begin{eqnarray*}
\bepsilon &=& \frac{{\cal T}^*_F  \times_1 \widehat{\bbeta}_{1} \times_2  \widehat{\bbeta}_{2}}{\|{\cal T}_F  \times_1 \widehat{\bbeta}_{1} \times_2  \widehat{\bbeta}_{2}\|_2} - \bbeta_3^* + \frac{{\cal E}_F  \times_1 \widehat{\bbeta}_{1} \times_2  \widehat{\bbeta}_{2}}{\|{\cal T}_F  \times_1 \widehat{\bbeta}_{1} \times_2  \widehat{\bbeta}_{2}\|_2}. 
\end{eqnarray*}
To bound $\|\bepsilon\|_{\infty}$, it is sufficient to bound the following two terms, $(I)$ and $(II)$, where 
$$
\change{\|\bepsilon\|_{\infty} \le \underbrace{\left\| \frac{{\cal T}^*_F  \times_1 \widehat{\bbeta}_{1} \times_2  \widehat{\bbeta}_{2}}{\|{\cal T}_F  \times_1 \widehat{\bbeta}_{1} \times_2  \widehat{\bbeta}_{2}\|_2} - \bbeta_3^* \right\|_{\infty} }_{(I)} + \underbrace{ \frac{ \| {\cal E}_F  \times_1 \widehat{\bbeta}_{1} \times_2  \widehat{\bbeta}_{2} \|_{\infty}}{\|{\cal T}_F  \times_1 \widehat{\bbeta}_{1} \times_2  \widehat{\bbeta}_{2}\|_2}}_{(II)}.}
$$

Before we compute the upper bound of $(I)$ and $(II)$, respectively, we first derive the lower bound of $\|{\cal T}_F  \times_1 \widehat{\bbeta}_{1} \times_2  \widehat{\bbeta}_{2}\|_2$. By the model assumption ${\cal T} = {\cal T}^* + {\cal E}$, we have
\begin{eqnarray}
\|{\cal T}_F  \times_1 \widehat{\bbeta}_{1} \times_2  \widehat{\bbeta}_{2}\|_2 &\ge& \|{\cal T}_F^*  \times_1 \widehat{\bbeta}_{1} \times_2  \widehat{\bbeta}_{2}\|_2 - \|{\cal E}_F  \times_1 \widehat{\bbeta}_{1} \times_2  \widehat{\bbeta}_{2}\|_2 \label{eqn:denominator_lowerbound1}\\
&\ge& w^* (1-\epsilon_0^2) - \eta({\cal E}; s_1, s_2, s_3), \label{eqn:denominator_lowerbound}
\end{eqnarray}
where the second inequality is due to the following two facts. First, $\|{\cal T}^*_F  \times_1 \widehat{\bbeta}_{1} \times_2  \widehat{\bbeta}_{2}\|_2 = \|{\cal T}^*  \times_1 \widehat{\bbeta}_{1} \times_2  \widehat{\bbeta}_{2}\|_2 = \|  w^* \langle \widehat{\bbeta}_{1}, \bbeta^*_{1} \rangle \langle \widehat{\bbeta}_{2}, \bbeta^*_{2} \rangle \bbeta_3^* \|_2 = w^* |\langle \widehat{\bbeta}_{1}, \bbeta^*_{1} \rangle | | \langle \widehat{\bbeta}_{2}, \bbeta^*_{2} \rangle | \ge w^* (1-\epsilon_0^2)$, where the last inequality holds because of the initial conditions on $\widehat{\bbeta}_{1}, \widehat{\bbeta}_{2}$, as well as the fact that $\langle \bu, \bv \rangle^2 \ge 1 - \| \bu - \bv \|_2^2$ for unit vectors $\bu,\bv$. Second, by definition of tensor norm and the property $\|\widehat{\bbeta}_{1}\|=\|\widehat{\bbeta}_{2}\| =1$, we have
\begin{eqnarray}
 \eta({\cal E}; s_1, s_2, s_3) &=&\sup_{\substack{\| \ub \| = \|\vb\| = \|\wb\| = 1\\ \| \ub \|_0 \le s_1, \|\vb\|_0 \le s_2, \|\wb\|_0 \le s_3}} \Big|{\cal E} \times_1 \ub \times_2 \vb \times_3 \wb \Big|  \nonumber \\
&\ge& \sup_{\| \ub \| = \|\vb\| = \|\wb\| = 1} \Big|{\cal E}_F \times_1 \ub \times_2 \vb \times_3 \wb \Big|  \ge  \|{\cal E}  \times_1 \widehat{\bbeta}_{1} \times_2  \widehat{\bbeta}_{2}\|_2, \label{eqn:error_spectral_bound}
\end{eqnarray}
where the last inequality is due to the Cauchy-Schwarz inequality. The lower bound in $(\ref{eqn:denominator_lowerbound})$ is positive according to the assumption on the initialization and error tensor.

Next we bound the two terms $(I)$ and $(II)$. \change{To bound $(I)$, we have that}
\begin{eqnarray}
&&\change{\frac{{\cal T}^*_F  \times_1 \widehat{\bbeta}_{1} \times_2  \widehat{\bbeta}_{2}}{\|{\cal T}_F  \times_1 \widehat{\bbeta}_{1} \times_2  \widehat{\bbeta}_{2}\|_2} - \bbeta_3^*}  \nonumber \\
&\le& \change{\frac{{\cal T}^*_F  \times_1 \widehat{\bbeta}_{1} \times_2  \widehat{\bbeta}_{2}  - \|{\cal T}^*_F  \times_1 \widehat{\bbeta}_{1} \times_2  \widehat{\bbeta}_{2}\|_2 \bbeta_3^* + \|{\cal E}_F  \times_1 \widehat{\bbeta}_{1} \times_2  \widehat{\bbeta}_{2}\|_2 \bbeta_3^*  }{\|{\cal T}_F  \times_1 \widehat{\bbeta}_{1} \times_2  \widehat{\bbeta}_{2}\|_2}}   \label{eqn:error_bound_1} \\
&=& \change{\frac{\|{\cal E}_F  \times_1 \widehat{\bbeta}_{1} \times_2  \widehat{\bbeta}_{2}\|_2 \bbeta_3^*  }{\|{\cal T}_F  \times_1 \widehat{\bbeta}_{1} \times_2  \widehat{\bbeta}_{2}\|_2}} \label{eqn:error_bound_2}\\
&\le& \change{ \frac{ \eta({\cal E}; s_1, s_2, s_3) \bbeta_3^* }{w^* (1-\epsilon_0^2) - \eta({\cal E}; s_1, s_2, s_3)}.} \label{eqn:error_bound_3}
\end{eqnarray}
\change{where the inequality in $(\ref{eqn:error_bound_1})$ is due to $(\ref{eqn:denominator_lowerbound1})$ and the condition that $(\ref{eqn:denominator_lowerbound})$ is positive, the equality in $(\ref{eqn:error_bound_2})$ is due to the above argument that $\|{\cal T}^*_F  \times_1 \widehat{\bbeta}_{1} \times_2  \widehat{\bbeta}_{2}\|_2 = w^* |\langle \widehat{\bbeta}_{1}, \bbeta^*_{1} \rangle | | \langle \widehat{\bbeta}_{2}, \bbeta^*_{2} \rangle | = w^* \langle \widehat{\bbeta}_{1}, \bbeta^*_{1} \rangle \langle \widehat{\bbeta}_{2}, \bbeta^*_{2} \rangle$. Here $\langle \widehat{\bbeta}_{i}, \bbeta^*_{i} \rangle > 0$ is due to the initialization condition on $\|\widehat{\bbeta}_{i} - \bbeta^*_{i}\|$ for $i=1,2$. Finally, the last inequality $(\ref{eqn:error_bound_3})$ is due to $(\ref{eqn:denominator_lowerbound})$ and $(\ref{eqn:error_spectral_bound})$. Therefore, this fact, together with $\|\bbeta_3^* \|_{\infty} \le 1$, implies that}
$$
(I) \le  \frac{ \eta({\cal E}; s_1, s_2, s_3)}{w^* (1-\epsilon_0^2) - \eta({\cal E}; s_1, s_2, s_3)}.
$$
In addition, according to $(\ref{eqn:error_spectral_bound})$, we have $\| {\cal E}  \times_1 \widehat{\bbeta}_{1} \times_2  \widehat{\bbeta}_{2} \|_{\infty} \le  \eta({\cal E}; s_1, s_2, s_3)$, and hence
$$
(II) \le  \frac{ \eta({\cal E}; s_1, s_2, s_3)}{w^* (1-\epsilon_0^2) - \eta({\cal E}; s_1, s_2, s_3)}.
$$
Therefore, we have the final bound for $\|\bepsilon\|_{\infty}$ in that, 
$$
\|\bepsilon\|_{\infty} \le \frac{ 2\eta({\cal E}; s_1, s_2, s_3)}{w^* (1-\epsilon_0^2) - \eta({\cal E}; s_1, s_2, s_3)}.
$$
The rest of the theorem follows from $(\ref{eqn:lemma_bound_beta})$ in Lemma \ref{lemma:local} and the assumption $\| \Db \bbeta_3^* \|_1 \le f_{03}$. This completes the proof of Theorem \ref{theorem:local}. \eop

\subsection{Proof of Corollary \ref{thm:final}}

According to Theorem \ref{theorem:local}, it remains to show the upper bound in $(\ref{eqn:bound_beta3})$ is bounded by $\epsilon_0^2/4$. Therefore, it suffices to show
\begin{equation}
\left[\frac{ 2\eta({\cal E}; s_1, s_2, s_3) }{w^*(1 - \epsilon_0^2) - \eta({\cal E}; s_1, s_2, s_3)}\right]^2 \le \frac{\epsilon_0^2}{8}, \;\; \textrm{ and } \;\; 
\frac{8M f_{0,3}  \eta({\cal E}; s_1, s_2, s_3)}{w^*(1 - \epsilon_0^2) - \eta({\cal E}; s_1, s_2, s_3)} \le \frac{\epsilon_0^2}{8}. \label{eqn:ensure_contract}
\end{equation}
Note that when $ \eta({\cal E}; s_1, s_2, s_3) \le w^* \epsilon_0 (1-\epsilon_0)/9$, we have, $\eta({\cal E}; s_1, s_2, s_3) \le \{w^* \epsilon_0 (1-\epsilon_0^2)\} / (8+\epsilon_0)$, 
due to the fact that $\epsilon_0 \le 1$. Therefore,
\begin{eqnarray*}
\change{\frac{ 2\eta({\cal E}; s_1, s_2, s_3) }{w^*(1 - \epsilon_0^2) - \eta({\cal E}; s_1, s_2, s_3)} \le \frac{2w^* \epsilon_0 (1-\epsilon_0^2)/ (8+\epsilon_0)}{ w^*(1-\epsilon_0^2)-w^* \epsilon_0 (1-\epsilon_0^2)/ (8+\epsilon_0)} = \epsilon_0/4,}
\end{eqnarray*}
which implies directly that the first argument in $(\ref{eqn:ensure_contract})$ holds.

Moreover, when $\eta({\cal E}; s_1, s_2, s_3) \le \change{w^*\epsilon_0^2(1-\epsilon_0)/(64 M f_{0,3} + 1)}$, we have
$$
\change{\eta({\cal E}; s_1, s_2, s_3) \le \frac{w^*\epsilon_0^2(1-\epsilon_0)(1+\epsilon_0)}{64 M f_{0,3} + 1} \le \frac{w^*\epsilon_0^2(1-\epsilon_0^2)}{64 M f_{0,3} + \epsilon_0^2}}.
$$
Therefore, we have
\begin{eqnarray*}
\change{\frac{8M f_{0,3}  \eta({\cal E}; s_1, s_2, s_3)}{w^*(1 - \epsilon_0^2) - \eta({\cal E}; s_1, s_2, s_3)}  \le \frac{8M f_{0,3}  w^*\epsilon_0^2(1-\epsilon_0^2)/(64 M f_{0,3} + \epsilon_0^2) }{w^*(1 - \epsilon_0^2) - w^*\epsilon_0^2(1-\epsilon_0^2)/(64 M f_{0,3} + \epsilon_0^2)} = \epsilon_0^2/8,}
\end{eqnarray*}
which validates the second argument in $(\ref{eqn:ensure_contract})$. This completes the proof of Corollary \ref{thm:final}. \eop

\subsection{Proof of Corollary \ref{cor:gaussian_error}}
\label{sec:proof_corollary_gaussian_error}

The derivation in the Gaussian error tensor scenario consists of three steps. In Step 1, we show via the large deviation bound inequality that, for some Gaussian tensor $\cG$,
$$
\mathbb P \left( |\eta\left(\cG; s_1, s_2, s_3\right) - \mathbb E[\eta\left(\cG; s_1, s_2, s_3\right)]| \ge t \right) \le 2\exp\left( -\frac{t^2}{2L^2}\right)
$$
with some Lipschitz constant $L$. In Step 2, by incorporating Lemma~\ref{lemma:gaussian_width}, and exploring the sparsity constraint, we establish that  
$$
\mathbb E[\eta\left(\cG; s_1, s_2, s_3\right)]| \le C \sqrt{s_1s_2s_3 \log(d_1d_2d_3)},
$$
for some constant $C>0$. In Step 3, we derive the final rate by incorporating the above results into Theorem \ref{theorem:local}.

Step 1: We show that the function $\eta\left(\cdot; s_1, s_2, s_3\right)$ is a Lipschitz function in its first argument. For any two tensors $\cG_1,\cG_2\in \mathbb R^{d_1 \times d_2 \times d_3}$, denote $\cA^* =  \sup_{\cA} \langle \cG_1,  \cA \rangle$. We have 
$$
\sup_{\cA} \langle \cG_1,  \cA \rangle - \sup_{\cA} \langle \cG_2,  \cA \rangle \le \langle \cG_1,  \cA^* \rangle - \sup_{\cA} \langle \cG_2,  \cA \rangle \le \langle \cG_1,  \cA^* \rangle - \langle \cG_2,  \cA^* \rangle \le \langle \cG_1 - \cG_2,  \cA^* \rangle.
$$  
Therefore, by the definition of $\eta\left(\cdot; s_1, s_2, s_3\right)$, we have 
\begin{eqnarray*}
|\eta\left(\cG_1; s_1, s_2, s_3\right) - \eta\left(\cG_2; s_1, s_2, s_3\right)| 
& \le &  \sup_{\substack{\| \bu \| = \|\bv\| = \|\bw\| = 1\\ \| \bu \|_0 \le s_1, \|\bv\|_0 \le s_2, \|\bw\|_0 \le s_3}} \Big\langle \cG_1 - \cG_2,  \bu \circ \bv  \circ\bw \Big\rangle\\
& \le & \sup_{\| \bu \| = \|\bv\| = \|\bw\| = 1} \| \bu \circ \bv  \circ \bw\|_F \cdot \|\cG_1 - \cG_2\|_F
\le \|\cG_1 - \cG_2\|_F,
\end{eqnarray*}
where the second inequality is due to the fact that $\langle \cA, \cB \rangle \le \|\cA\|_F\|\cB\|_F$, and the third inequality is due to $ \| \bu \circ \bv  \circ \bw \|_F =  \| \bu \circ \bv  \circ \bw \|_2 \le \|\bu\|_2\|\bv\|_2\|\bw\|_2 = 1$ for any unit-norm vectors $\bu, \bv, \bw$.

Applying the concentration result of Lipschitz functions of Gaussian random variables in Lemma \ref{lemma:concentration} with $L=1$, for the Gaussian tensor $\cG$, we have
\begin{equation}
\mathbb P \left(|\eta\left(\cG; s_1, s_2, s_3\right) - \mathbb E[\eta\left(\cG; s_1, s_2, s_3\right)]| \ge t \right) \le 2\exp\left( -\frac{t^2}{2}\right).
\label{eqn:stage2}
\end{equation}

Step 2: We aim to bound $\mathbb E[\eta\left(\cG ; s_1, s_2, s_3\right)]$. For a tensor ${\cal T}\in \mathbb R^{d_1 \times d_2 \times d_3}$, denote its $L_1$-norm regularizer as $R({\cal T}) := \sum_{j_1}\sum_{j_2}\sum_{j_3 } |{\cal T}_{j_1,j_2,j_3}|$. Define the ball of this regularizer as $B_{R}(\delta) := \{{\cal T}\in \mathbb R^{d_1 \times d_2 \times d_3}| R({\cal T}) \le \delta \}$. For any vectors $\bu \in \mathbb R^{d_1}, \bv  \in \mathbb R^{d_2}, \bw  \in \mathbb R^{d_3}$ satisfying $\| \bu \|_2 = \|\bv\|_2 = \|\bw\|_2 = 1, \| \bu \|_0 \le s_1, \|\bv\|_0 \le s_2$, and $\|\bw\|_0 \le s_3$, denote $\cA:=  \bu \circ \bv \circ \bw$. Lemma \ref{lemma:L1normbound} implies that $R({\cal A}) \le \sqrt{s_1s_2s_3}$. Therefore, we have, 
\begin{eqnarray*}
\mathbb E[\eta\left(\cG ; s_1, s_2, s_3\right)] &=& \mathbb E\left[  \sup_{\substack{\| \bu \| = \|\bv\| = \|\bw\| = 1\\ \| \bu \|_0 \le s_1, \|\bv\|_0 \le s_2, \|\bw\|_0 \le s_3}} \langle \cG, \bu \circ \bv  \circ \bw \rangle \right]\\
&\le&  \mathbb E\left[  \sup_{\cA \in B_{R}(\sqrt{s_1s_2s_3})} \langle \cG, \cA \rangle\right]  = \sqrt{s_1s_2s_3}\mathbb \; \mathbb E\left[  \sup_{\cA \in B_{R}(1)} \langle \cG, \cA \rangle\right].
\end{eqnarray*}
This result, together with Lemma \ref{lemma:gaussian_width}, implies that
\begin{equation}
\mathbb E[\eta\left(\cG ; s_1, s_2, s_3\right)] \le C \sqrt{s_1s_2s_3 \log(d_1d_2d_3)}.
\label{eqn:stage3}
\end{equation}

Finally, combing $(\ref{eqn:stage2})$ and $(\ref{eqn:stage3})$, and setting $t =  \sqrt{s_1s_2s_3 \log(d_1d_2d_3)}$, we have, with probability $1-2\exp( -s_1s_2s_3 \log(d_1d_2d_3)/2)$, 
$$
|\eta\left(\cG ; s_1, s_2, s_3\right) - \mathbb E[\eta\left(\cG ; s_1, s_2, s_3\right)]| \le  \sqrt{s_1s_2s_3 \log(d_1d_2d_3)}. 
$$
Henceforth,
$$
\eta\left(\cG ; s_1, s_2, s_3\right) \le (C+1)\sqrt{s_1s_2s_3 \log(d_1d_2d_3)}.
$$

Step 3: Finally, we derive the rate of $\widehat{\bbeta}_{3}$ by incorporating the inequality of $\eta(\cG ; s_1, s_2, s_3)$ in Stage 2 into the general rate in Theorem \ref{theorem:local}. Here for notational simplicity, we denote the one-step estimator $\widehat{\bbeta}_{3}^{(1)}$ in our algorithm as $\widehat{\bbeta}_{3}$. In particular, when $w^* \succ \sqrt{s_1s_2s_3 \log(d_1d_2d_3)}$, there exists a constant $0<c<1$, such that $w^*(1 - \epsilon_0^2) - (C+1)\sqrt{s_1s_2s_3 \log(d_1d_2d_3)} \ge c w^*(1 - \epsilon_0^2)$. Therefore,
$$
\| \widehat{\bbeta}_{3} - \bbeta_3^*\|^2_2 \le \left[\frac{ 2(C+1)\sqrt{s_1s_2s_3 \log(d_1d_2d_3)} }{cw^*(1 - \epsilon_0^2)}\right]^2 + \frac{8M f_{0,3}  (C+1)\sqrt{s_1s_2s_3 \log(d_1d_2d_3)} }{cw^*(1 - \epsilon_0^2)},
$$
and henceforth, up to a logarithm term, we have, 
$$
\| \widehat{\bbeta}_{3} - \bbeta_3^*\|^2_2 = \tilde{O}_p\left( \max \left\{ \frac{s_1s_2s_3}{w^{*2}}, \frac{f_{0,3} \sqrt{s_1s_2s_3}}{w^*}   \right\}  \right).
$$
This completes the proof of Corollary \ref{cor:gaussian_error}. \eop

\subsection{Proof of Theorem \ref{thm:cluster}}
\label{sec:proof_thm_clusterr}

As in Theorem \ref{theorem:local}, for notational simplicity, we denote the one-step estimator $\widehat{\bbeta}_{3}^{(1)}$ in our algorithm as $\widehat{\bbeta}_{3}$.  Our proof consists of three steps. In Step 1, we derive the upper bound of $\|D \bbeta_3^*\|_1$ given the clustering assumption in $(\ref{eqn:beta3_cluster})$. In Step 2, we use the general theory developed in Corollary \ref{cor:gaussian_error} to obtain the final rate. In Step 3, we derive the clustering consistency via the rate of convergence in Stage 2. 

Step 1: Recall that in $(\ref{eqn:beta3_cluster})$, we assume the component $\bbeta_3^* \in \mathbb R^N$ has the following clustering structure, 
$$
\bbeta_3^* = ( \underbrace{ \mu^*_{1},\ldots,  \mu^*_{1}}_{l \textrm{~elements}}, \; \underbrace{ \mu^*_{2}, \ldots, \mu^*_{2}}_{l \textrm{~elements}}, \; \ldots, \; \underbrace{\mu^*_{K}, \ldots, \mu^*_{K}}_{l \textrm{~elements}}),
$$
To facilitate the derivation, we denote $\bbeta := \bbeta_3^*$ and denote its $j$th entry as $\beta_j$. Then we have, 
$$
\|D \bbeta\|_1 = \sum_{j=2}^{N} |\beta_{j} - \beta_{j-1}|  = \sum_{j=2}^{K} |\mu^*_j - \mu^*_{j-1}| \le 2 \sum_{j=1}^K |\mu^*_j| \le 2 \sqrt{K}  \sqrt{ \sum_{j=1}^K \mu^{*2}_j}
$$
where the second equality holds because of the above structural assumption of $\bbeta_3^*$, and the last inequality holds due to the Cauchy-Schwarz inequality. Moreover, due to the unit-norm condition, we have $\|\bbeta\|_2 = 1$; that is, $\sum_{j=1}^K \mu^{*2}_j N/K = 1$. Therefore, we have $\sum_{j=1}^K \mu^{*2}_j = K/N$, and 
\begin{equation}
\|D \bbeta_3^*\|_1 = \|D \bbeta\|_1 \le \frac{2K}{\sqrt{N}}.
\label{eqn:bound_fuse}
\end{equation}

Step 2: Note that $\bbeta_3^*$ is not necessarily sparse and hence $\|\bbeta_3^*\|_0 \le s_3 = N$. According to the general theory developed in Corollary \ref{cor:gaussian_error}, we have
$$
\| \widehat{\bbeta}_{3} - \bbeta_3^*\|^2_2 \le \left[\frac{C_1\sqrt{s_1s_2N \log(d_1d_2N)} }{w^*}\right]^2 + \frac{C_2 f_{0,3} \sqrt{s_1s_2N \log(d_1d_2N)} }{w^*},
$$
for some constants $C_1$ and $C_2$. Here $f_{0,3}$ is the upper bound of $\|D \bbeta_3^*\|_1$, and is bounded by $2K/\sqrt{N}$ according to $(\ref{eqn:bound_fuse})$. Therefore, when $w^* \succ \sqrt{s_1s_2N^2 \log(d_1d_2N)/K}$, we have
$$
\left[\frac{C_1\sqrt{s_1s_2N \log(d_1d_2N)} }{w^*}\right]^2 = \change{O\left(\frac{K}{N}\right)}, \;\; \textrm{~and~} \;\; \frac{C_2 f_{0,3} \sqrt{s_1s_2N \log(d_1d_2N)} }{w^*} = O\left(\frac{K^2}{N}\right),
$$
and henceforth, 
$$
\| \widehat{\bbeta}_{3} - \bbeta_3^*\|^2_2 = O\left(\frac{K^2}{N}\right).
$$

Step 3: Based on the rate in Stage 2, we have $\| \widehat{\bbeta}_{3} - \bbeta_3^*\|_{\infty} \le \| \widehat{\bbeta}_{3} - \bbeta_3^*\|_2 = O(K/\sqrt{N}),$
where $\|\bbeta\|_{\infty} = \max_j{\beta_j}$. Therefore, when the minimal gap between two clusters is lower bounded such that $\min_{i, j} |\mu^*_{i} - \mu^*_{j}| > C_1K/\sqrt{N}$, it is guaranteed to recover all cluster structures. That is, with high probability, we have $\widehat{\cA}_k = \cA_k^*,$ for any $k$. This completes the proof of Theorem \ref{thm:cluster}. \eop

{\color{black}
\subsection{Proof of Theorem \ref{theorem:local_general}} 

For notational simplicity, we denote the one-step estimator $\widehat{\bbeta}_{3,r}^{(1)}$ as $\widehat{\bbeta}_{3,r}$, and the initial estimators $\widehat{\bbeta}_{1,r}^{(0)}, \widehat{\bbeta}_{2,r}^{(0)}$ as $\widehat{\bbeta}_{1,r}, \widehat{\bbeta}_{2,r}$, respectively. Denote $F_1 := \textrm{supp}(\bbeta_{1,r}^*) \cup \textrm{supp}(\widehat{\bbeta}_{1,r})$, $F_2 := \textrm{supp}(\bbeta_{2,r}^*) \cup \textrm{supp}(\widehat{\bbeta}_{2,r})$, and $F_3 := \textrm{supp}(\bbeta_{3,r}^*) \cup \textrm{supp}(\widehat{\bbeta}_{3,r})$, and let $F := F_1 \circ F_2 \circ F_3$. Following similar arguments as that in the proof of Theorem \ref{theorem:local}, the key step is to compute $\|\bepsilon\|_{\infty}$, where
\begin{eqnarray*}
\bepsilon &=& \textrm{Norm}\big({\cal T}_F \times_1 \widehat{\bbeta}_{1,r} \times_2  \widehat{\bbeta}_{2,r}   \big) - \bbeta_{3,r}^*\\
&=& \underbrace{\frac{{\cal T}^*_F  \times_1 \widehat{\bbeta}_{1,r} \times_2  \widehat{\bbeta}_{2,r}}{\|{\cal T}_F  \times_1 \widehat{\bbeta}_{1,r} \times_2  \widehat{\bbeta}_{2,r}\|_2} - \bbeta_{3,r}^*}_{(I)} + \underbrace{\frac{{\cal E}_F  \times_1 \widehat{\bbeta}_{1,r} \times_2  \widehat{\bbeta}_{2,r}}{\|{\cal T}_F  \times_1 \widehat{\bbeta}_{1,r} \times_2  \widehat{\bbeta}_{2,r}\|_2}}_{(II)}. 
\end{eqnarray*}
\change{Next, to bound $\|\bepsilon\|_{\infty}$, we divide our procedure in three steps. In Step 1, we decompose and bound the nominator ${\cal T}^*_F  \times_1 \widehat{\bbeta}_{1,r} \times_2  \widehat{\bbeta}_{2,r}$ in the term $(I)$. In Step 2, we seek the lower bound of the denominator $\|{\cal T}_F  \times_1 \widehat{\bbeta}_{1,r} \times_2  \widehat{\bbeta}_{2,r}\|_2$ in $(I)$. In Step 3, we bound $\|(I)\|_{\infty}$ and $\|(II)\|_{\infty}$, respectively, then eventually bound $\|\bepsilon\|_{\infty}$. }

\change{Step 1: For the nominator ${\cal T}^*_F  \times_1 \widehat{\bbeta}_{1,r} \times_2  \widehat{\bbeta}_{2,r}$ in $(I)$,} consider $\widehat{\bbeta}_{1,r} = \widehat{\bbeta}_{1,r} - \bbeta_{1,r}^* + \bbeta_{1,r}^*$ and $\widehat{\bbeta}_{2,r} = \widehat{\bbeta}_{2,r} - \bbeta_{2,r}^* + \bbeta_{2,r}^*$. We have
\begin{eqnarray}
{\cal T}^*_F  \times_1 \widehat{\bbeta}_{1,r} \times_2  \widehat{\bbeta}_{2,r} &=&  \underbrace{{\cal T}^*_F  \times_1 (\widehat{\bbeta}_{1,r} - \bbeta_{1,r}^*) \times_2  (\widehat{\bbeta}_{2,r}- \bbeta_{2,r}^*)}_{I_1} + \underbrace{{\cal T}^*_F  \times_1 (\widehat{\bbeta}_{1,r} - \bbeta_{1,r}^*) \times_2  \bbeta_{2,r}^*}_{I_2} \nonumber\\
&& +  \underbrace{{\cal T}^*_F  \times_1 \bbeta_{1,r}^* \times_2  (\widehat{\bbeta}_{2,r}- \bbeta_{2,r}^*)}_{I_3} +  \underbrace{{\cal T}^*_F  \times_1 \bbeta_{1,r}^* \times_2  \bbeta_{2,r}^*}_{I_4} \label{eqn:bound_numeric}
\end{eqnarray}
According to Lemma \ref{lemma:pre2} and the CP decomposition in $(\ref{eqn:CP})$, we have
$$
{\cal T}^*_F = \sum_{j\in [R]} w_j^* \textrm{Truncate}(\bbeta_{1,j}^*, {F_1}) \circ \textrm{Truncate}(\bbeta_{2,j}^*, {F_2}) \circ \textrm{Truncate}(\bbeta_{3,j}^*, {F_3}).
$$
Denote $\bar{\bbeta}_{i,j}^* = \textrm{Truncate}(\bbeta_{i,j}^*, {F_i})$ for $i=1,2,3$ and $j=1,\ldots,R$. We have
$$
I_1 = \sum_{j\in [R]} w_j^* \langle \bar{\bbeta}_{1,j}^*, \widehat{\bbeta}_{1,r} - \bbeta_{1,r}^* \rangle \langle \bar{\bbeta}_{2,j}^*, \widehat{\bbeta}_{2,r} - \bbeta_{2,r}^* \rangle  \bar{\bbeta}_{3,j}^*.
$$
This, together with the Cauchy-Schwarz inequality $|\langle \bbeta_1, \bbeta_2 \rangle| \le \|\bbeta_1\|_2 \|\bbeta_2\|_2$, implies that
\begin{eqnarray}
\| I_1 \|_{\infty} &\le& \sum_{j\in [R]} w_j^* | \langle \bar{\bbeta}_{1,j}^*, \widehat{\bbeta}_{1,r} - \bbeta_{1,r}^* \rangle| | \langle \bar{\bbeta}_{2,j}^*, \widehat{\bbeta}_{2,r} - \bbeta_{2,r}^* \rangle | \| \bar{\bbeta}_{3,j}^*\|_{\infty} \nonumber \\
&\le& \| \widehat{\bbeta}_{1,r} - \bbeta_{1,r}^*\|_2  \| \widehat{\bbeta}_{2,r} - \bbeta_{2,r}^*\|_2  \sum_{j\in [R]} w_j^* \| \bar{\bbeta}_{1,j}^* \|_2 \| \bar{\bbeta}_{2,j}^*\|_2 \| \bar{\bbeta}_{3,j}^*\|_{\infty} \nonumber \\
&\le& \epsilon_0^2  \sum_{j\in [R]} w_j^* \le \epsilon_0^2 \| {\cal T}^*\| \le C_1 w_{\max} \epsilon_0^2  \label{eqn:bound_I1},
\end{eqnarray}
where the third inequality is due to the fact that $\| \bar{\bbeta}_{i,j}^* \|_2 \le \| \bbeta_{i,j}^* \|_2 = 1$ for $i=1,2$, and $\| \bar{\bbeta}_{3,j}^*\|_{\infty} = \| \bbeta_{3,j}^*\|_{\infty} \le 1$. The forth inequality is due to the property of the spectral norm of the true tensor. In particular, $\|{\cal T}^*\| = \sup_{\|\ub\|=\|\vb\|=\|\wb\|=1} |{\cal T}^* \times_1 \ub \times_2 \vb \times_3 \wb| = \sup_{\|\ub\|=\|\vb\|=\|\wb\|=1} |\sum_{j\in [R]} w_j^* \langle \ub, \bbeta_{1,j}^* \rangle  \langle \vb, \bbeta_{2,j}^* \rangle \langle \wb, \bbeta_{3,j}^* \rangle | \ge \sum_{j\in [R]} w_j^* $ by letting $\ub = \bbeta_{1,j}^*, \vb = \bbeta_{2,j}^*, \wb = \bbeta_{3,j}^*$. Finally, the last inequality is \change{due to Assumption \ref{ass:modelr1}.}

Next, we have
\begin{eqnarray*}
I_2 &=&  \sum_{j\in [R]} w_j^* \langle \bar{\bbeta}_{1,j}^*, \widehat{\bbeta}_{1,r} - \bbeta_{1,r}^* \rangle \langle \bar{\bbeta}_{2,j}^*, \bbeta_{2,r}^* \rangle  \bar{\bbeta}_{3,j}^*\\
&=& \underbrace{w_r^* \langle \bbeta_{1,r}^*, \widehat{\bbeta}_{1,r} - \bbeta_{1,r}^* \rangle  \bbeta_{3,r}^*}_{I_{21}} +    \underbrace{\sum_{j\ne r} w_j^* \langle \bar{\bbeta}_{1,j}^*, \widehat{\bbeta}_{1,r} - \bbeta_{1,r}^* \rangle \langle \bar{\bbeta}_{2,j}^*, \bbeta_{2,r}^* \rangle  \bar{\bbeta}_{3,j}^*}_{I_{22}}, 
\end{eqnarray*}
where the second equality is due to the fact that $\langle \bar{\bbeta}_{1,j}^*, \widehat{\bbeta}_{1,r} - \bbeta_{1,r}^* \rangle = \langle \bbeta_{1,r}^*, \widehat{\bbeta}_{1,r} - \bbeta_{1,r}^* \rangle$ since $\textrm{supp}(\widehat{\bbeta}_{1,r}) \subseteq F_1$ and $\textrm{supp}(\widehat{\bbeta}_{1,r} - \bbeta_{1,r}^*) \subseteq F_1$, as well as the fact $\langle \bar{\bbeta}_{2,r}^*, \bbeta_{2,r}^* \rangle = \langle \bbeta_{2,r}^*, \bbeta_{2,r}^* \rangle =1$. By the Cauchy-Schwarz inequality, we have $\|I_{21}\|_{\infty} \le w_r^* \|  \bbeta_{1,r}^*\|_2\| \widehat{\bbeta}_{1,r} - \bbeta_{1,r}^*\|_2 \|  \bbeta_{3,r}^*\|_{\infty} \le w_r^*\epsilon_0 $. Moreover, 
\begin{eqnarray}
\|I_{22}\|_{\infty} &\le& \sum_{j\ne r} w_j^*\| \bar{\bbeta}_{1,j}^*\|_2 \| \widehat{\bbeta}_{1,r} - \bbeta_{1,r}^*\|_2  \langle \bar{\bbeta}_{2,j}^*, \bbeta_{2,r}^* \rangle \| \bar{\bbeta}_{3,j}^*\|_{\infty} \nonumber  \\
&\le& \sum_{j\ne r} w_j^* \epsilon_0 \xi \le \epsilon_0 \xi (R-1) w_{\max}, \label{eqn:bound_I22}
\end{eqnarray}
where the second inequality is due to the facts that $\| \bar{\bbeta}_{1,j}^*\|_2 \le 1$, $\| \bar{\bbeta}_{3,j}^*\|_{\infty} \le 1$, and $\langle \bar{\bbeta}_{2,j}^*, \bbeta_{2,r}^* \rangle \le \langle \bbeta_{2,j}^*, \bbeta_{2,r}^* \rangle \le \xi$ with the incoherence parameter $\xi$ defined in $(\ref{eqn:incoherence})$. Therefore, we have
\begin{equation*}
\| I_2 \|_{\infty} \le \|I_{21}\|_{\infty} + \|I_{22}\|_{\infty} \le w_r^*\epsilon_0 + \epsilon_0 \xi (R-1) w_{\max} \label{eqn:bound_I2}.
\end{equation*}
By a similar argument, we also have
\begin{equation*}
\| I_3 \|_{\infty} \le w_r^*\epsilon_0 + \epsilon_0 \xi (R-1) w_{\max} \label{eqn:bound_I3}.
\end{equation*}

Finally, for $I_4$, we have
\begin{eqnarray}
I_4 &=&  \sum_{j\in [R]} w_j^* \langle \bar{\bbeta}_{1,j}^*, \bbeta_{1,r}^* \rangle \langle \bar{\bbeta}_{2,j}^*, \bbeta_{2,r}^* \rangle  \bar{\bbeta}_{3,j}^* \nonumber \\
&=& w_r^* \bbeta_{3,r}^* +    \underbrace{\sum_{j\ne r} w_j^* \langle \bar{\bbeta}_{1,j}^*, \bbeta_{1,r}^* \rangle \langle \bar{\bbeta}_{2,j}^*, \bbeta_{2,r}^* \rangle  \bar{\bbeta}_{3,j}^*}_{I_{42}}. \label{eqn:bound_I4}
\end{eqnarray}
According to the Cauchy-Schwarz inequality, and similar arguments used in the bound of $\|I_{22}\|_{\infty}$, we have
\begin{equation}
\|I_{42}\|_{\infty} \le \sum_{j\ne r} w_j^*| \langle \bar{\bbeta}_{1,j}^*, \bbeta_{1,r}^* \rangle ||\langle \bar{\bbeta}_{2,j}^*, \bbeta_{2,r}^* \rangle | \|\bar{\bbeta}_{3,j}^*\|_{\infty} \le \xi^2 (R-1)w_{\max}. \label{eqn:bound_I42}
\end{equation}

\change{Step 2: Denote the denominator in the term $(I)$ as $\alpha := \|{\cal T}_F  \times_1 \widehat{\bbeta}_{1,r} \times_2  \widehat{\bbeta}_{2,r}\|_2$. We next derive the lower bound of $\alpha$.} According to $(\ref{eqn:CP})$, and the facts $\textrm{supp}(\widehat{\bbeta}_{1,r}) \subseteq F_1$, $\textrm{supp}(\widehat{\bbeta}_{2,r}) \subseteq F_2$, and $\textrm{supp}({\bbeta}^*_{3,r}) \subseteq F_3$, we have
\begin{eqnarray*}
&& {\cal T}_F  \times_1 \widehat{\bbeta}_{1,r} \times_2  \widehat{\bbeta}_{2,r} = {\cal T}^*_F  \times_1 \widehat{\bbeta}_{1,r} \times_2  \widehat{\bbeta}_{2,r} + {\cal E}_F  \times_1 \widehat{\bbeta}_{1,r} \times_2  \widehat{\bbeta}_{2,r}\\
&=&   \underbrace{w_r^* \langle \bbeta_{1,r}^*, \widehat{\bbeta}_{1,r} \rangle \langle \bbeta_{2,r}^*, \widehat{\bbeta}_{2,r} \rangle \bbeta_{3,r}^*}_{I_5} +    \underbrace{\sum_{j\ne r} w_j^* \langle \bbeta_{1,j}^*, \widehat{\bbeta}_{1,r} \rangle \langle \bbeta_{2,j}^*, \widehat{\bbeta}_{2,r} \rangle  \bar{\bbeta}_{3,j}^*}_{I_6} +  \underbrace{ {\cal E}_F  \times_1 \widehat{\bbeta}_{1,r} \times_2  \widehat{\bbeta}_{2,r}}_{I_7}.
\end{eqnarray*}
This implies that
$$
\alpha = \|{\cal T}_F  \times_1 \widehat{\bbeta}_{1,r} \times_2  \widehat{\bbeta}_{2,r}\|_2 \ge \|I_5\|_2 - \|I_6\|_2 - \|I_7\|_2.
$$
Therefore, in order to find the lower bound of $\alpha$, it is sufficient to find the lower bound of $\|I_5\|_2$ and upper bounds of $\|I_6\|_2$ and $\|I_7\|_2$. The initial conditions imply that $\|I_5\|_2 \ge w_r^*(1- \epsilon_0^2)$. Moreover, according to the argument in $(\ref{eqn:error_spectral_bound})$, we have $\|I_7\|_2 = \|{\cal E}_F  \times_1 \widehat{\bbeta}_{1,r} \times_2  \widehat{\bbeta}_{2,r}\|_2 \le \|{\cal E}_F\| \le \eta({\cal E}; s_1, s_2, s_3)$. Furthermore, considering $\widehat{\bbeta}_{1,r} = \widehat{\bbeta}_{1,r} - \bbeta_{1,r}^* + \bbeta_{1,r}^*$ and $\widehat{\bbeta}_{2,r} = \widehat{\bbeta}_{2,r} - \bbeta_{2,r}^* + \bbeta_{2,r}^*$, we have
\begin{eqnarray*}
&& \|I_6\|_2 \\
&\le&  \underbrace{\|\sum_{j\ne r} w_j^* \langle \bbeta_{1,j}^*, \widehat{\bbeta}_{1,r} - \bbeta_{1,r}^* \rangle \langle \bbeta_{2,j}^*, {\bbeta}^*_{2,r} \rangle  \bar{\bbeta}_{3,j}^*\|_2}_{I_{61}} + \underbrace{\|\sum_{j\ne r} w_j^* \langle \bbeta_{1,j}^*, {\bbeta}^*_{1,r} \rangle \langle \bbeta_{2,j}^*, \widehat{\bbeta}_{2,r} - \bbeta_{2,r}^* \rangle  \bar{\bbeta}_{3,j}^*\|_2}_{I_{62}} \\
&+& \underbrace{\|\sum_{j\ne r} w_j^* \langle \bbeta_{1,j}^*, \widehat{\bbeta}_{1,r} - \bbeta_{1,r}^* \rangle \langle \bbeta_{2,j}^*, \widehat{\bbeta}_{2,r} - \bbeta_{2,r}^* \rangle  \bar{\bbeta}_{3,j}^*\|_2}_{I_{63}} + \underbrace{\|\sum_{j\ne r} w_j^* \langle \bbeta_{1,j}^*, \bbeta^*_{1,r} \rangle \langle \bbeta_{2,j}^*, \bbeta_{2,r}^* \rangle  \bar{\bbeta}_{3,j}^*\|_2}_{I_{64}}
\end{eqnarray*}
Using similar arguments as those in $(\ref{eqn:bound_I1})$, $(\ref{eqn:bound_I22})$, and $(\ref{eqn:bound_I42})$, we have
$$
I_{61} \le \epsilon_0 \xi (R-1) w_{\max}, \; I_{62} \le \epsilon_0 \xi (R-1) w_{\max}, \; I_{63} \le C_1\epsilon_0^2 w_{\max},  \; I_{64} \le \xi^2(R-1) w_{\max},
$$
and hence the denominator $\alpha := \|{\cal T}_F  \times_1 \widehat{\bbeta}_{1,r} \times_2  \widehat{\bbeta}_{2,r}\|_2$ satisfies 
\begin{equation}
\alpha\ge w_r^*(1- \epsilon_0^2) -  w_{\max} \underbrace{\left[ 2\epsilon_0 \xi (R-1) + C_1\epsilon_0^2  + \xi^2(R-1) \right]}_{g(\epsilon_0, \xi, R)} - \eta({\cal E}; s_1, s_2, s_3) \label{eqn:bound_alpha}
\end{equation}

\change{Step 3:} Combining the results in $(\ref{eqn:bound_numeric})$, $(\ref{eqn:bound_I4})$ and $(\ref{eqn:bound_alpha})$, we have
\begin{eqnarray*}
(I) &=& \frac{{\cal T}^*_F  \times_1 \widehat{\bbeta}_{1,r} \times_2  \widehat{\bbeta}_{2,r} - \alpha \bbeta_{3,r}^*}{\alpha} \\
&=&  \frac{I_1 + I_2 + I_3 + w_r^* \bbeta_{3,r}^* + I_{42}- \alpha \bbeta_{3,r}^*}{\alpha} \\
&\le&  \frac{I_1 + I_2 + I_3 + I_{42} + w_r^* \bbeta_{3,r}^* - \left[ w_r^*(1- \epsilon_0^2) -  w_{\max} g(\epsilon_0, \xi, R) - \eta({\cal E}; s_1, s_2, s_3)  \right] \bbeta_{3,r}^*}{w_r^*(1- \epsilon_0^2) -  w_{\max} g(\epsilon_0, \xi, R) - \eta({\cal E}; s_1, s_2, s_3) }
\end{eqnarray*}
This, together with the bounds on $I_1$,  $I_2$,  $I_3$, and $I_{42}$, implies that
$$
\|(I)\|_{\infty} \le  \frac{ 2w_{\max} g(\epsilon_0, \xi, R) + \eta({\cal E}; s_1, s_2, s_3)}{w_r^*(1- \epsilon_0^2) -  w_{\max} g(\epsilon_0, \xi, R) - \eta({\cal E}; s_1, s_2, s_3)}
$$

Following similar arguments as those in $(\ref{eqn:error_spectral_bound})$ and $(\ref{eqn:bound_alpha})$, we have
$$
\|(II)\|_{\infty} \le  \frac{ \eta({\cal E}; s_1, s_2, s_3)}{w_r^*(1- \epsilon_0^2) -  w_{\max} g(\epsilon_0, \xi, R) - \eta({\cal E}; s_1, s_2, s_3)}.
$$
Therefore, we have
$$
\|\bepsilon\|_{\infty} \le  \frac{ 2w_{\max} g(\epsilon_0, \xi, R) + 2\eta({\cal E}; s_1, s_2, s_3)}{w_r^*(1- \epsilon_0^2) -  w_{\max} g(\epsilon_0, \xi, R) - \eta({\cal E}; s_1, s_2, s_3)}.
$$
The rest of the theorem follows from $(\ref{eqn:lemma_bound_beta})$ in Lemma \ref{lemma:local} and the conditions on $\lambda_3$ and $\| \Db \bbeta_{3,r}^* \|_1$. This completes the proof of Theorem \ref{theorem:local_general}. \eop

\subsection{Proof of Corollary \ref{cor:local_general_iteration}} 

Based on the results in Theorem \ref{theorem:local_general}, we prove Corollary \ref{cor:local_general_iteration} in two steps. In step 1, we find the lower bound of the denominator $w_r^*(1- \epsilon_0^2) -  w_{\max} g(\epsilon_0, \xi, R) - \eta({\cal E}; s_1, s_2, s_3)$ in $(\ref{eqn:bound_beta3_general})$. In Step 2, we simplify its numerator and find its upper bound. 

Step 1: According to the first condition in Assumption \ref{ass:initial}, we have $\tilde{q}:= C_1 \epsilon_0 + 2 \xi (R-1) < w_{\min}/(8w_{\max}) <1$, which, together with the second condition in Assumption \ref{ass:initial}, implies that
$$
g(\epsilon_0, \xi, R) = \epsilon_0^2 C_1 + 2 \epsilon_0 \xi (R-1) + \xi^2 (R-1) = \xi^2 (R-1) + \tilde{q}\epsilon_0 \le \frac{w_{\min}}{6w_{\max}}.
$$
Moreover, the second condition in Assumption \ref{ass:initial} implies that $\epsilon_0 \le w_{\min} / 6w_{\max}$, and henceforth  
$\epsilon_0^2  w_{\max} / w_{\min} \le 1/6$. These two facts, together with Assumption \ref{ass:noise}, lead to the lower bound
\begin{eqnarray}
&&w_r^*(1- \epsilon_0^2) -  w_{\max} g(\epsilon_0, \xi, R) - \eta({\cal E}; s_1, s_2, s_3) \nonumber \\
&\ge& w_{\min} \left \{ 1 -   \frac{w_{\max}}{w_{\min}} \epsilon_0^2 -  \frac{w_{\max}}{w_{\min}} g(\epsilon_0, \xi, R) - \frac{ \eta({\cal E}; s_1, s_2, s_3)}{w_{\min}}\right\} \nonumber\\
&>&  w_{\min}(1- \frac{1}{6}- \frac{1}{6}- \frac{1}{6}) = \frac{w_{\min}}{2}. \label{eqn:bound_halfwmin}
\end{eqnarray}

Step 2: According to Theorem \ref{theorem:local_general}, $(\ref{eqn:bound_halfwmin})$ and the Assumption \ref{ass:fuse}, we have
\begin{eqnarray*}
\| \widehat{\bbeta}_{3,r}^{(1)} - \bbeta_{3,r}^*\|_2 &\le& \frac{8w_{\max}}{w_{\min}} g(\epsilon_0, \xi, R) +  \frac{8}{w_{\min}} \eta({\cal E}; s_1, s_2, s_3)\\
&\le&  \frac{8w_{\max}}{w_{\min}}[ \xi^2 (R-1) + \tilde{q}\epsilon_0]+  \frac{8}{w_{\min}} \eta({\cal E}; s_1, s_2, s_3)\\
&\le&  \frac{8w_{\max}\tilde{q}}{w_{\min}} \epsilon_0 +  \frac{8w_{\max}}{w_{\min}}\xi^2 (R-1) + \frac{8}{w_{\min}} \eta({\cal E}; s_1, s_2, s_3)\\
& = & q \epsilon_0 + \epsilon_S,
\end{eqnarray*}
where the contraction coefficient $q := 8w_{\max}\tilde{q} / w_{\min} < 1$ by the first condition in Assumption \ref{ass:initial}, and the statistical error $\epsilon_S$ is independent of the initial error $\epsilon_0$. The rest of Corollary \ref{cor:local_general_iteration} follows by iteratively applying the above inequality. This completes the proof of Corollary \ref{cor:local_general_iteration}. \eop

\subsection{Proof of Theorem \ref{thm:cluster_general}} 

Based on the results in Corollary \ref{cor:local_general_iteration}, the estimator $\widehat{\bbeta}_{m+1,r}$ after $T$ iterations   satisfies that 
$$
\max_{r = 1,\ldots, R} \| \widehat{\bbeta}_{m+1,r} - \bbeta_{m+1,r}^*\|_2 \le O_p(\epsilon_S).
$$
According to Corollary \ref{cor:gaussian_error}, when the error tensor is a Gaussian tensor, we have that $\eta({\cal E}; s_1, \ldots, s_{m+1}) = C\sqrt{\prod_{j=1}^{m+1} s_j \sum_{j=1}^{m+1}\log(d_j)}$ for some constant $C$. Note that here $s_{m+1} = d_{m+1} = N$. Therefore,
$$
\epsilon_S \sim \frac{w_{\max}}{w_{\min}}\xi^2 (R-1) + \frac{\sqrt{N\log(N)\prod_{j=1}^{m} s_j \sum_{j=1}^{m}\log(d_j)}}{w_{\min}} 
$$
where $a \sim b$ means $a,b$ are in the same order. Therefore, when $w_{\max} / w_{\min} \le C_2$ for some constant $C_2>0$, $\xi^2(R-1) = O(K/\sqrt{N})$, and $w_{\min} \succ \sqrt{\prod_{j=1}^{m} s_j N^2 \log(\prod_{j=1}^{m} d_{j} N)/K}$, we have $\epsilon_S= O( K / \sqrt{N})$. Based on this result, we obtain the estimation error in clustering; i.e.,
$$
\max_{i} \|\hat{\bmu}_i - \bmu^*_i \|_2 \le \sqrt{R} \max_{r = 1,\ldots, R} \| \widehat{\bbeta}_{m+1,r} - \bbeta_{m+1,r}^*\|_2 \le O_p\left(K\sqrt{\frac{R}{N}}\right) < \tilde{C}K\sqrt{R/N},
$$
for some constant $\tilde{C}$.

Finally, if $\min_{i, j} \|\bmu^*_{i} - \bmu^*_{j}\|_2 > C_3K\sqrt{R/N}$ for some constant $C_3 > 4\tilde{C}$, we have, for any two samples $\hat{\bmu}_i, \hat{\bmu}_j$ from different clusters $\cA_i^*$ and $\cA_j^*$, respectively,
\begin{eqnarray*}
\|\hat{\bmu}_i - \hat{\bmu}_j \|_2 &=& \|\hat{\bmu}_i - \bmu^*_{i} + \bmu^*_{i} - \bmu^*_{j} +\bmu^*_{j} - \hat{\bmu}_j \|_2 \\
&\ge& \|\bmu^*_{i} - \bmu^*_{j}\|_2 - \|\hat{\bmu}_i - \bmu^*_{i}\|_2 - \|\bmu^*_{j} - \hat{\bmu}_j \|_2 \\
&>& 2\tilde{C}K\sqrt{R/N},
\end{eqnarray*}
Similarly, for any two samples $\hat{\bmu}_i, \hat{\bmu}_{i'}$ from the same cluster $\cA_i^*$, 
\begin{eqnarray*}
\|\hat{\bmu}_i - \hat{\bmu}_{i'} \|_2 &=& \|\hat{\bmu}_i - \bmu^*_{i} + \bmu^*_{i} - \hat{\bmu}_{i'} \|_2 \\
&\le& \|\hat{\bmu}_i - \bmu^*_{i}\|_2 - \|\bmu^*_{i} - \hat{\bmu}_{i'} \|_2 \\
&\le& 2\tilde{C}K\sqrt{R/N}.
\end{eqnarray*}
This implies that the within-cluster distance is always smaller than the between-cluster distance, and henceforth, any Euclidean-distance based clustering algorithm is able to correctly assign the clusters of all $\hat{\bmu}_i$, $i=1,\ldots, N$. That is, we have $\widehat{\cA}_k = \cA_k^*$ for any $k=1,\ldots,K$, with high probability. This completes the proof of Theorem \ref{thm:cluster_general}. \eop

\subsection{Additional numerical analyses} 
\label{sec:additional_experiments}

In this section, we present additional simulations with different correlation structures, with large ranks, and with unequal cluster sizes. We also report the clustering analysis in the first task of our real data analysis but in the frequency domain instead of the time domain.

\begin{table}[t!]
\centering
\begin{small}
\caption{Clustering of 2D matrices with different correlation structures. The methods under comparison are the same as described in  Table \ref{tab:sim_2d}.}
\label{tab:add_sim_2d}
\vskip 1 em
\begin{tabular}{ccc|cccc} \hline
 & & & \multicolumn{4}{c}{Tensor recovery error} \\ \cline{4-7}
  Correlation & $N$ & $\rho$ & STF & TTP & GLTD \\ \hline
AR&50 & 0.1 & \textbf{0.317} (0.024) & 0.446 (0.034) & 0.550 (0.040) \\ 
  & & 0.2 & \textbf{0.448} (0.035) & 0.516 (0.035) & 0.645 (0.036) \\ 
  & & 0.5 & \textbf{0.699} (0.038) & 0.710 (0.039) & 0.972 (0.037) \\ 
  &100  & 0.1 & \textbf{0.277} (0.022) & 0.405 (0.035) & 0.479 (0.038) \\ 
  && 0.2 & \textbf{0.334} (0.033) & 0.413 (0.036) & 0.520 (0.040) \\ 
  & & 0.5 & \textbf{0.655} (0.028) & 0.678 (0.026) & 0.923 (0.032) \\ 
Exchangeable& 50  & 0.1 & \textbf{0.464} (0.035) & 0.512 (0.036) & 0.814 (0.034) \\ 
  &  & 0.2 & 0.909 (0.041) & \textbf{0.883} (0.04) & 1.010 (0.032) \\ 
  &  & 0.5 & 1.946 (0.034) & 1.913 (0.037) & \textbf{1.878} (0.025) \\ 
  &100  & 0.1 & \textbf{0.356} (0.029) & 0.407 (0.035) & 0.721 (0.043) \\ 
  &  & 0.2 & \textbf{0.880} (0.046) & 0.898 (0.044) & 1.061 (0.023) \\ 
  & & 0.5 & 1.998 (0.031) & 1.966 (0.032) & \textbf{1.890} (0.023) \\ 
\hline
& & & \multicolumn{4}{c}{Clustering error} \\ \cline{4-7}
 Correlation &$N$ & $\rho$ & DTC & TTP & GLTD & vectorized \\ 
  \hline
AR&50 &  0.1 & \textbf{0.036} (0.013) & 0.092 (0.017) & 0.148 (0.023) & 0.255 (0.000) \\ 
  & & 0.2 & \textbf{0.117} (0.018) & 0.148 (0.018) & 0.201 (0.021) & 0.255 (0.000) \\ 
  & & 0.5 & \textbf{0.237} (0.02) & 0.250 (0.023) & 0.381 (0.023) & 0.256 (0.001) \\ 
  &100  & 0.1 & \textbf{0.025} (0.011) & 0.086 (0.017) & 0.113 (0.019) & 0.253 (0.000) \\ 
  & & 0.2 & \textbf{0.066} (0.016) & 0.092 (0.017) & 0.146 (0.021) & 0.253 (0.000) \\ 
  & & 0.5 & \textbf{0.214} (0.015) & 0.227 (0.013) & 0.363 (0.019) & 0.253 (0.000) \\ 
Exchangeable&  50 & 0.1 &\textbf{0.124} (0.020) & 0.142 (0.020) & 0.270 (0.021) & 0.255 (0.000) \\ 
  & & 0.2 & 0.248 (0.024) & \textbf{0.236} (0.023) & 0.265 (0.023) & 0.309 (0.012) \\ 
  & & 0.5 & \textbf{0.272} (0.013) & 0.281 (0.017) & 0.290 (0.018) & 0.491 (0.005) \\ 
  &100  & 0.1 & \textbf{0.063} (0.015) & 0.086 (0.017) & 0.217 (0.022) & 0.253 (0.000) \\ 
  & & 0.2 & \textbf{0.226} (0.024) & 0.238 (0.023) & 0.283 (0.018) & 0.288 (0.012) \\ 
  & & 0.5 & 0.275 (0.014) & 0.267 (0.014) & \textbf{0.264} (0.016) & 0.486 (0.005) \\ 
\hline
\end{tabular}
\end{small}
\end{table}

First, we investigate the impact of different correlation structures on our method. We adopt the same simulation setting as in Section \ref{sec:sim-2d}, but replace the original identity covariance matrix with a covariance matrix under an autoregressive (AR) structure or an exchangeable structure. That is, we generate the data from \eqref{eqn:general_data_generation}, with $\bSigma_{k,j}$, $k=1,\ldots,K, j=1,\ldots,m$, of the form, 

\begin{gather}
\bSigma_{k,j}^{\text{ar}}  =
\begin{pmatrix} 
1 & \rho & \cdots \rho^{d_j-1} & \rho^{d_j} \\
\rho & 1 & \cdots \rho^{d_j-2}  & \rho^{d_j-1}  \\
\vdots & \vdots & \vdots & \vdots \\
\rho^{d_j} & \rho^{d_j-1}&   \cdots \rho & 1 \\
\end{pmatrix} \in \mathbb R^{d_j \times d_j}, \nonumber \quad
\bSigma_{k,j}^{\text{exch}}  =
\begin{pmatrix} 
1 & \rho & \rho  \cdots & \rho \\
\rho & 1 & \rho  \cdots & \rho  \\
\vdots & \vdots & \vdots & \vdots\\
\rho & \rho & \rho  \cdots& 1 \\
\end{pmatrix} \in \mathbb R^{d_j \times d_j}. \nonumber
\end{gather}
We consider the correlation coefficient $\rho \in \{0.1, 0.2, 0.5\}$, where a larger value of $\rho$ indicates a more severe misspecification for our method. We vary the sample size $N\in \{50, 100\}$, and fix the other parameters $\mu = 1.2$ and $d_1=d_2=20$. Table \ref{tab:add_sim_2d} reports the tensor recovery error and clustering error out of 50 data replications for all methods under the two correlation structures. It is seen that, when the correlation is moderate, i.e., $\rho = 0.1, 0.2$, our method clearly outperforms the alternative solutions in terms of both tensor recovery and clustering accuracy. When the correlation increases to $0.5$, for which our model is severely misspecified, our method still works reasonable well. In addition, the computational time under the AR or the  exchangeable correlation structure is very similar to that reported in Table \ref{tab:sim_2d} under the independent correlation structure, and we omit the results.

Next we investigate the impact of a large rank $R$ on our method. We continue to adopt the simulation setting as in Section \ref{sec:sim-2d}, but increase the true rank value $R\in \{5,8\}$. We vary the sample size $N\in \{50, 100\}$, $\mu \in \{1, 1.2\}$, and fix $d_1 = d_2 = 20$. Table \ref{tab:largeRerror} reports the tensor recovery error and clustering error out of 50 data replications for all methods under the new rank values. We continue to observe that our method performs competitively with a larger value of $R$.

\begin{table}[t!]
\centering
\begin{small}
\caption{Clustering of 2D matrices with different ranks. The methods under comparison are the same as described in Table \ref{tab:sim_2d}.}
\label{tab:largeRerror}
\vskip 1 em
\begin{tabular}{cccc|cccc} \hline
& & & & \multicolumn{4}{c}{Tensor recovery error} \\ \cline{5-8}
  $d_1 = d_2$ & $N$ & $\mu$ & Rank & STF & TTP & GLTD \\ \hline
20 &50 & 1 & 5 & \textbf{0.764} (0.032) & 0.848 (0.036) & 0.865 (0.036) \\ 
 &   &  & 8 & \textbf{0.808} (0.021) & 0.862 (0.021) & 0.903 (0.023) \\ 
 &  & 1.2 & 5 & \textbf{0.509} (0.007) & 0.589 (0.030) & 0.609 (0.031) \\ 
 &  &  & 8 & \textbf{0.668} (0.006) & 0.689 (0.013) & 0.728 (0.022) \\ 
 & 100 & 1 & 5 & \textbf{0.712} (0.032) & 0.748 (0.034) & 0.831 (0.034) \\ 
 &  &  & 8 & \textbf{0.804} (0.018) & 0.849 (0.021) & 0.930 (0.022) \\ 
 &  & 1.2 & 5 & \textbf{0.496} (0.004) & 0.572 (0.028) & 0.659 (0.034) \\ 
 &  &  & 8 & \textbf{0.640} (0.011) & 0.665 (0.020) & 0.678 (0.017) \\ 
  \hline
&& & & \multicolumn{4}{c}{Clustering error} \\ \cline{5-8}
 $d_1 = d_2$ &$N$ & $\mu$ & Rank & DTC & TTP & GLTD & vectorized \\ 
  \hline
20 &50 & 1 & 5 & \textbf{0.183} (0.029) & 0.259 (0.035) & 0.245 (0.036) & 0.291 (0.007) \\ 
&   &  & 8 & \textbf{0.147} (0.031) & 0.222 (0.031) & 0.247 (0.033) & 0.280 (0.005) \\ 
&   & 1.2 & 5 & \textbf{0.013} (0.013) & 0.047 (0.019) & 0.064 (0.024) & 0.256 (0.001) \\ 
&   &  & 8 & \textbf{0.000} (0.000) & 0.033 (0.019) & 0.066 (0.026) & 0.255 (0.000) \\ 
&  100 & 1 & 5 & \textbf{0.139} (0.032) & 0.156 (0.034) & 0.224 (0.032) & 0.278 (0.004) \\ 
&  &  & 8 & \textbf{0.165} (0.027) & 0.200 (0.028) & 0.314 (0.032) & 0.266 (0.003) \\ 
&  & 1.2 & 5 & \textbf{0.000} (0.000) & 0.040 (0.017) & 0.115 (0.027) & 0.234 (0.014) \\ 
&  & & 8 & \textbf{0.013} (0.013) & 0.025 (0.023) & 0.050 (0.021) & 0.038 (0.021) \\ 
\hline
\end{tabular}
\end{small}
\end{table}

Next we consider a simulation example where the cluster sizes are unequal. We adopt the simulation setting in Section \ref{sec:sim-3d}, but set the four cluster sizes $l_1,\ldots,l_4$ satisfying that $l_1: l_2:l_3:l_4 = 1:2:3:4$. Table \ref{tab:sim_3d_unequal} reports the tensor recovery error and clustering error out of 50 data replications. We observe a similar pattern of performance when the cluster sizes are different. In this article, the assumption of equal cluster sizes is imposed only to simplify the presentation. It can be easily relaxed to the general case of unequal cluster sizes. 

\begin{table}[t!]
\centering
\begin{small}
\caption{Clustering of 3D tensors with unequal cluster sizes. The methods under comparison are the same as described in Table \ref{tab:sim_2d}.}
\label{tab:sim_3d_unequal}
\vskip 1 em
\begin{tabular}{ccc|cccc} \hline
& & & \multicolumn{4}{c}{Tensor recovery error} \\ \cline{4-7}
 $d_1 = d_2 = d_3$ & $N$ & $\mu$ & STF & TTP & GLTD \\ \hline
20 & 50 & 0.6 & \textbf{0.052} (0.003) & 0.699 (0.049) & 0.346 (0.081)\\ 
 &   & 0.8 & \textbf{0.012} (0.001) & 0.335 (0.070) & 0.119 (0.048) \\ 
  & 100 & 0.6 & \textbf{0.038} (0.003) & 0.797 (0.047) & 0.316 (0.053)\\ 
 &  & 0.8 & \textbf{0.009} (0.001) & 0.237 (0.059) & 0.066 (0.037) \\ \hline
 & & & \multicolumn{4}{c}{Clustering error} \\ \cline{4-7}
  $d_1 = d_2 = d_3$ &$N$ & $\mu$ & DTC & TTP & GLTD & vectorized \\ 
\hline
20 &  50 & 0.6 & \textbf{0.177} (0.028) & 0.400 (0.028) & 0.286 (0.035) & 0.406 (0.017) \\ 
 &    & 0.8 & \textbf{0.000} (0.000) & 0.146 (0.034) & 0.058 (0.026) & 0.204 (0.000)  \\ 
  & 100 & 0.6 & \textbf{0.069} (0.025) & 0.408 (0.031) & 0.191 (0.032) & 0.287 (0.008) \\ 
 &  & 0.8 & \textbf{0.000} (0.000) & 0.107 (0.031) & 0.029 (0.020) & 0.202 (0.000) \\ 
\hline
\end{tabular}
\end{small}
\end{table}

Finally, we add an analysis of the ABIDE fMRI data in the frequency domain to mitigate the potential issue of temporal inconsistency. Specifically, we repeat the first task in Section \ref{sec:realdata}, but transform the data to the frequency domain via fast Fourier transform (FFT). That is, following \citet{calhoun2003, ahn2015}, we calculate the temporal amplitude spectrum for each voxel, 
$X_{ik}  = | FFT(x_{ik}) |$,  
where $x_{ik}$ represents the original measure in the time domain at voxel $i$ and time point $k$. The rest of analysis is the same as the one in the time domain. Table \ref{tab:USM_error_FFT} reports the clustering error of all methods in the frequency domain. It is seen that our proposed method and the method of \citet{MS2016} perform similarly and are the best across all sliding windows. The smallest clustering error in the frequency domain is $21/57$, which is larger than the smallest clustering error $15/57$ obtained in the time domain in Table \ref{tab:USM_error}. 

\begin{table}[b!]
\centering
\begin{small}
\caption{Clustering of the ABIDE data along the \emph{subject mode} in the frequency domain. The setup is the same as Table \ref{tab:USM_error}.}
\label{tab:USM_error_FFT}
\vskip 1 em
\begin{tabular}{c|cccc} \hline
Windows & DTC & TTP & GLTD & vectorized \\
\hline
1 & 21/57 &   26/57 &   21/57 &   21/57  \\
30 & 21/57 &   25/57 &   21/57 &   21/57  \\
50 & 22/57 &   24/57 &   22/57 &   22/57  \\
80 & 21/57 &   22/57 &   21/57 &   24/57  \\
\hline
\end{tabular}
\end{small}
\end{table}

}

\end{document}